\documentclass[10pt]{article}
\usepackage{fullpage,graphicx,subfigure,mathdots,mathpazo,color}
\usepackage{amsmath,amscd,tikz,mathrsfs,cite}
\usepackage[normalem]{ulem}
\usepackage{amsmath}
\usepackage{setspace}

\usepackage{epsfig,amsmath,graphicx,amssymb,overpic}
\usepackage{bm}
\usepackage{graphicx}
\usepackage{subfigure, xcolor}
\usepackage{ntheorem}
\usepackage{listings,diagbox}

\usepackage{color}
\usepackage{epsfig,amsmath,graphicx,amssymb,overpic}
\usepackage{colortbl}

\usepackage{booktabs}
\usepackage{graphicx}
\usepackage{dcolumn}
\usepackage{bm}
\usepackage{subfigure}
\usepackage{epsfig,amsmath,graphicx,amssymb,overpic,cite}
\usepackage{makecell}
\usepackage{booktabs}
\usepackage{bm}
\usepackage{subfigure, xcolor}
\usepackage{float}
\usepackage{tabularx}
\usepackage{multirow}

\usepackage{algorithm}
\usepackage{algorithmic}

\def\be{\begin{equation}}
\def\ee{\end{equation}}
\def\bee{\begin{eqnarray}}
\def\ene{\end{eqnarray}}
\def\bes{\begin{subequations}}
\def\ees{\end{subequations}}

\newcommand{\bx}{{\bm x}}

\newcommand{\bu}{{\bm u}}
\newcommand{\bU}{{\bm U}}

\def\v{\vspace{0.06in}}

\def\be{\begin{equation}}
\def\ee{\end{equation}}
\def\bee{\begin{eqnarray}}
\def\ene{\end{eqnarray}}
\def\bes{\begin{subequations}}
\def\ees{\end{subequations}}

\usepackage{geometry}
\allowdisplaybreaks[4]

\begin{document}

\baselineskip=14pt
\renewcommand {\thefootnote}{\dag}
\renewcommand {\thefootnote}{\ddag}
\renewcommand {\thefootnote}{ }

\pagestyle{plain}

\begin{center}
\baselineskip=16pt \leftline{} \vspace{-.3in} {\Large \bf PMNO: A novel physics guided multi-step neural operator predictor for  partial differential equations} \\[0.2in]
\end{center}

\begin{center}
{\bf Jin Song}$^{a,b}$,\, {\bf Kenji Kawaguchi}$^{c}$,\, {\bf Zhenya Yan}$^{d,b,e,*}$
\footnote{$^*$ Corresponding author.\\
{\it Email addresses}: songjin21@mails.ucas.ac.cn (J. Song), kenji@nus.edu.sg (K. Kawaguchi), zyyan@mmrc.iss.ac.cn (Z. Yan).\\
} \\[0.15in]
{\it \small\baselineskip=13pt $^a$School of Advanced Interdisciplinary Science, University of Chinese Academy of Sciences, Beijing 100190, China\\
$^b$State Key Laboratory of Mathematical Sciences, Academy of Mathematics and Systems Science,\\
Chinese Academy of Sciences, Beijing 100190, China\\
$^c$National University of Singapore, 21 Lower Kent Ridge Road, 119077, Singapore\\
$^d$School of Mathematics and Information Science, Zhongyuan University of Technology, Zhengzhou 450007, China \\
$^e$School of Mathematical Sciences, University of Chinese Academy of Sciences, Beijing 100049, China
}
\end{center}

\vspace{0.1in} {\bf Abstract.}\, Neural operators, which aim to approximate mappings between infinite-dimensional function spaces, have been widely applied in the simulation and prediction of physical systems. However, the limited representational capacity of network architectures, combined with their heavy reliance on large-scale data, often hinder effective training and result in poor extrapolation performance. In  this paper, inspired by traditional numerical methods, we propose a novel physics guided multi-step neural operator (PMNO) architecture to address these challenges in long-horizon prediction of complex physical systems. Distinct from general operator learning methods, the PMNO framework replaces the single-step input with multi-step historical data in the forward pass and introduces an implicit time-stepping scheme based on the Backward Differentiation Formula (BDF) during backpropagation. This design not only strengthens the model's extrapolation capacity but also facilitates more efficient and stable training with fewer data samples, especially for long-term predictions.
Meanwhile, a causal training strategy is employed to circumvent the need for multi-stage training and to ensure efficient end-to-end optimization.
The neural operator architecture possesses resolution-invariant properties, enabling the trained model to perform fast extrapolation on arbitrary spatial resolutions.
We demonstrate the superior predictive performance of PMNO predictor across a diverse range of physical systems, including 2D linear system, modeling over irregular domain,
complex-valued wave dynamics, and reaction-diffusion processes.
Depending on the specific problem setting, various neural operator architectures, including FNO, DeepONet, and their variants, can be seamlessly integrated into the PMNO framework.

\vspace{0.1in}   {\it Keywords:} Partial differential equations,\, Neural operators,\, Machine learning,\, linear multi-step scheme,\, backward differentiation formula,\, physics guided,\, causal training




\vspace{0.2in}


\baselineskip=14pt

\section{Introduction}\label{sec1}

Spatiotemporal dynamics (e.g., chaotic dynamics, soliton interactions, and pattern formation) lies at the heart of physical systems \cite{physys1,physys2,physys3}, and plays a crucial role across a wide range of scientific and engineering domains, including fluid mechanics, climate science, optics, Bose-Einstein condensates, biology, and finance \cite{app-optic,app-bec,app-nrp,app-ocean,app-climate,app-financial}. Predicting and simulating their complex spatiotemporal dynamics from the available physical information is vital for controlling wave propagations, as well as for exploring the fundamental principles governing wave interactions in nonlinear media.

Current approaches generally follow two divergent paradigms: physics-based numerical discretization and purely data-driven surrogate modeling.
Fristly, traditional numerical solvers are primarily built upon time-integration schemes \cite{BDF,yjk,num-ana} coupled with spatial discretization techniques \cite{spectral,fdm,fem}.
Grounded in first principles, these methods offer high numerical accuracy, but they still suffer from prohibitive computational costs when resolving fine-scale features in high-dimensional systems, a challenge exacerbated by the ``curse of dimensionality".
Conversely, recent advances in machine learning have enabled data-driven strategies for modeling and understanding the behavior of complex physical systems \cite{pinn0,pinn1,ewn,ml0,ml1,RC,convlstm,transformer,transformer-nn,pdenet}.
Such successes are largely driven by the expressive power of deep neural networks, allowing them to learn intricate nonlinear dependencies from rich supervised data \cite{deeplearning}.
For example, machine learning techniques like convolutional neural networks (CNNs) and recurrent architectures bypass explicit physics encoding by learning input-output mappings directly from observational or synthetic datasets \cite{pdenet,convlstm}.
Moreover, Transformer architectures leverage self-attention mechanisms to capture dependencies across different positions within a sequence, making them particularly effective for modeling long-horizon temporal dynamics \cite{transformer,transformer-nn}.
While such black-box models demonstrate impressive acceleration in some tasks like flow field reconstruction, their reliance on extensive training data and opaque extrapolation mechanisms raises concerns about generalizability to unseen operating regimes \cite{theory}.
Purely data-driven methods based on deep learning typically extract representations from large datasets and heavily rely on data availability, which often leads to poor generalization and failure to adhere to underlying physical laws.
Moreover, the reliance on fixed-resolution data and positional encodings limits the model’s flexibility and effectiveness in capturing multi-scale features or dealing with irregular domains.

To address these limitations, neural operators have emerged as a promising class of deep learning models that aim to learn mappings between infinite-dimensional function spaces, circumventing spatial discretization constraints inherent to conventional PDEs solvers.
Pioneering works like DeepONet and Fourier Neural Operators (FNO) demonstrate orders-of-magnitude speedup in predicting fluid dynamics and solid mechanics, and have been supported by universal approximation theorem \cite{deeponet,fno,universal-deeponet,universal-fno,universal-fno1}.
Moreover, subsequent developments have introduced a variety of operator learning frameworks tailored to different problem settings. For instance, Graph Kernel Network (GNO) \cite{GKN} are designed for learning over unstructured meshes or irregular geometries, while Multiwavelet-based operators \cite{WNO0,WNO} aim to enhance the efficiency and accuracy in multi-scale problems.
These operator learning methods have also been applied to the prediction of various physical models and have achieved promising results \cite{NO-rnn,LNO,deeponet-predict,fno-predict}.
Inspired by the physics-informed learning paradigm introduced by PINNs, physics-embedded neural operator models, such as physics-informed DeepONet (PIdeepONet) and PIFNO \cite{PI-fno,PI-deeponet}, have been proposed to alleviate the reliance on large labeled datasets. These methods incorporate prior knowledge of governing physical laws to enable efficient learning in small-data regimes.
However, the design of most existing neural operator architectures remains largely intuition-driven, which often limits their representational capacity when applied to complex physical systems. Moreover, the incorporation of physical constraints, while beneficial for data efficiency, can significantly complicate the optimization landscape, making it challenging for the network to make long-horizon prediction.
These challenges call for a robust modeling framework that can harness the strengths of operator learning while mitigating its inherent limitations.

Motivated by the aforementioned discussions, in this paper we put forward the physics guided multi-step neural operator (PMNO) method, a novel neural operator architecture that incorporates the advantages of linear multi-step methods to address the challenge in long-horizon prediction of complex physical systems.
Unlike conventional operator learning approaches, the PMNO framework leverages multi-step historical data in the forward pass instead of relying on single-step inputs, and incorporates an implicit time-stepping mechanism rooted in the Backward Differentiation Formula (BDF) during the training process. Our main contributions of this paper can be summarized as follows:
\begin{itemize}
\item The multi-step neural operator structures enhance the temporal dependency across different time steps, thereby improving the network's expressive capacity for capturing complex dynamical behaviors.

\item BDF guided training paradigm enhances the model’s extrapolation capability, and facilitates more efficient and stable training with limited data. And the causal training strategy is employed to circumvent the need for multi-stage training and to ensure efficient end-to-end optimization.

\item Despite its architectural simplicity, PMNO predictor performs fast extrapolation on arbitrary spatial resolutions, and
achieves superior predictive performance across a diverse range of physical systems, including
2D linear system, modeling over irregular domain, complex-valued wave dynamics, and reaction-diffusion processes.

\end{itemize}

The remainder of this paper is arranged as follows. In Sec.~2, we formulate the main research problems, and introduce the related works and existing challenges.
Then in Sec.~3, we propose the framework and algorithm of the PMNO method, and present several practical techniques to enhance implementation efficiency and model performance, including hard-constraint boundary conditions and the causal training strategy.
In Sec.~4, we demonstrate the superior predictive capabilities of PMNO predictor by experimenting with a wide range of physical systems.
We also provide the corresponding algorithm analysis in Sec.~5.  Finally, some conclusions and discussions are given in Sec.~6.
The Appendix includes the introduction of the main deep operator network, the details of BDF, and the description of network structures, hyperparameters and model parameters.

\section{Preliminaries}

\subsection{\it Problem statement}

In this paper, we focus on future time prediction of physical systems, which can be described by a fundamental class of time-dependent partial differential equations (PDEs) with initial conditions (ICs) and boundary conditions (BCs),
\begin{equation}\label{pde}
\left\{\begin{array}{ll}
  \bu_t=\mathcal{N}[\bx, t; \bu(\bx,t)],&\quad \bx\in\Omega,\,\, t\in\mathcal{T},\v\\
 \bu(\bx,0)=\bu_0(\bx), &\quad \bx\in\Omega,\v\\
  \mathcal{B}[\bx,t;\bu(\bx,t)]=0,& \quad \bx\in\partial\Omega,\,\, t\in\mathcal{T},
 \end{array}\right.
\end{equation}
where $\bx\in \Omega\subseteq\mathbb{R}^d$ indicates the $d$-dimensional spatial coordinates, $\mathcal{T}=[0,+\infty)$ is the temporal domain, $\bu\in\mathcal{U}:\Omega\times\mathcal{T}\rightarrow\mathbb{R}^D$, and $\mathcal{N}:\Omega\times\mathcal{T}\times\mathcal{U}\rightarrow\mathbb{R}^D$ is a linear or nonlinear differential operator. $\bu_0\in\mathcal{U}$ is the initial condition, and $\mathcal{B}:\partial\Omega\times\mathcal{T}\times\mathcal{U}\rightarrow\mathbb{R}^D$ represents the boundary operator. For example, $\mathcal{B}[\bx,t;\bu(\bx,t)]=\bu(\bx,t)-g(\bx)$ indicates Dirichlet BCs with boundary value function $g(\bx)$, or $\mathcal{B}[\bx,t;\bu(\bx,t)]=\frac{\partial \bu}{\partial \vec{n}}$ represents Neumann BCs with unit outward normal vector $\vec{n}$ on $\partial\Omega$. The problems we consider are well-posed for Eq.~(\ref{pde}) under appropriate ICs and BCs.

In general, we aim to predict the future dynamical behaviors of the system based on the information of known solutions $\bU = [\bu_0, \bu_1, \ldots, \bu_{L-1}]$ at earlier time steps, where $\bu_i=\bu(\bx, i \Delta t)$ with discrete time interval $\Delta t$. In other words, we intend to design a predictor $\mathcal{P}$ such that $\bu_{L} =\mathcal{P}(\bU) $, which can consistently predict future behavior through a recurrent structure. Note that the predictor $\mathcal{P}$ is independent of time and space, and depends solely on the input states.

\subsection{\it  Related works and existing challenges}

A substantial body of research has been devoted to the simulation of physical systems. On the one hand, employing time-marching schemes and spatial discretization techniques constitutes the mainstream of traditional numerical methods.
In such methods, time-marching schemes, such as Euler method, Operator-splitting method, and Runge-Kutta method \cite{BDF,yjk,num-ana}, serve as the core iterative mechanism, progressing through discretized temporal intervals. And spectral methods, finite difference and finite element methods \cite{spectral,fdm,fem}, are employed as the spatial discretization schemes.

On the other hand, data-driven methods based on deep learning have recently attracted growing interest as an alternative to traditional simulation techniques.
One class of methods employs recurrent neural network (RNN) architectures for sequence prediction, such as reservoir computing and ConvLSTM \cite{RC,convlstm}.
In addition, Transformer models are capable of correlating different positions in a sequence to compute dependency representations via self-attention mechanisms, enabling them to effectively model long-horizon predictions \cite{transformer,transformer-nn}. Recently, several neural operators have been proposed to directly learn the evolution dynamics in the functional space of partial differential equations \cite{deeponet,fno}. These operator learning methods have also been applied to the prediction of various physical models and have achieved promising results \cite{NO-rnn,LNO,deeponet-predict,fno-predict}.

\v \textit{Challenges.}---Despite recent developments in simulation of physical systems, the field still faces several fundamental challenges, such as high-resolution results, handling high-dimensional data, scarce data problem,
effective long-horizon predictions, and maintaining a balance between computational cost and model expressiveness. On the one hand, traditional numerical methods, while strictly following to the governing equations, suffer from high computational overhead in extrapolation process especially in high dimension \cite{pinn1,ewn}. And the fixed resolution data and positional encodings tend to restrict its capacity when handling multi-scale and irregular domains problems. Likewise, RNN and Transformer structures face similar limitations, as they operate on data in discrete temporal and fixed spatial resolutions \cite{RC,convlstm,transformer-nn}.
On the other hand, data-driven approaches can offer fast extrapolation ability and partially mitigate the curse of dimensionality. And neural operators are capable of generalizing to inputs at arbitrary resolutions, as they learn mappings between infinite-dimensional function spaces.
However, data-driven models often demand substantial amounts of training data, while inferring the underlying physical dynamics from scarce data remains a significant challenge \cite{deeponet,fno}.
Besides, long-horizon forecasting typically presents optimization challenges, largely attributed to the inherent complexity of the models involved \cite{long-predict}.
These challenges call for a robust modeling framework that can harness the strengths of operator learning while mitigating its inherent limitations.

\section{Physics guided multi-step neural operator method}

Based on the above discussions, we propose the physics guided multi-step neural operator method (PMNO) to address the problem of long-horizon prediction in physical systems.
By leveraging the concept of linear multi-step scheme in numerical methods, we extend the input layer of the operator to incorporate more historical information, enabling long-horizon forecasting.
Guided by the prior physical knowledge, the model strictly follows to the governing equations, becoming more interpretable and less vulnerable to imperfect training data.
Furthermore, implicit time integration methods are considered to improve the stability of model for large time steps, thereby substantially lowering the training cost.
Meanwhile, hard-constraint boundary conditions and causal training techniques are also integrated into the models to accelerate the training process.

\subsection{\it Methodology}

For time interval $\Delta t$, we consider the general $k$-th order time-integrator for Eq.~(\ref{pde}) in the form of \cite{num-ana},
\begin{equation}\label{bash}
  \sum_{j=0}^{k}a_j\bu_{i+j} = \Delta t \sum_{j=0}^{k}b_j\mathcal{N}[\bx,t_{i+j};\bu_{i+j}],\quad i=0,1,\ldots,
\end{equation}
where $\bu_{i+j}=\bu(\bx, t_{i+j})$ with $t_{i+j} = (i+j)\Delta t$, $a_j$ and $b_j$ are the known parameters determined by the integration method, and $a_0\neq0$.
If $b_0=0$, the method becomes explicit and vice versa.

\begin{figure}[!t]
    \centering
      {\scalebox{0.65}[0.65]{\includegraphics{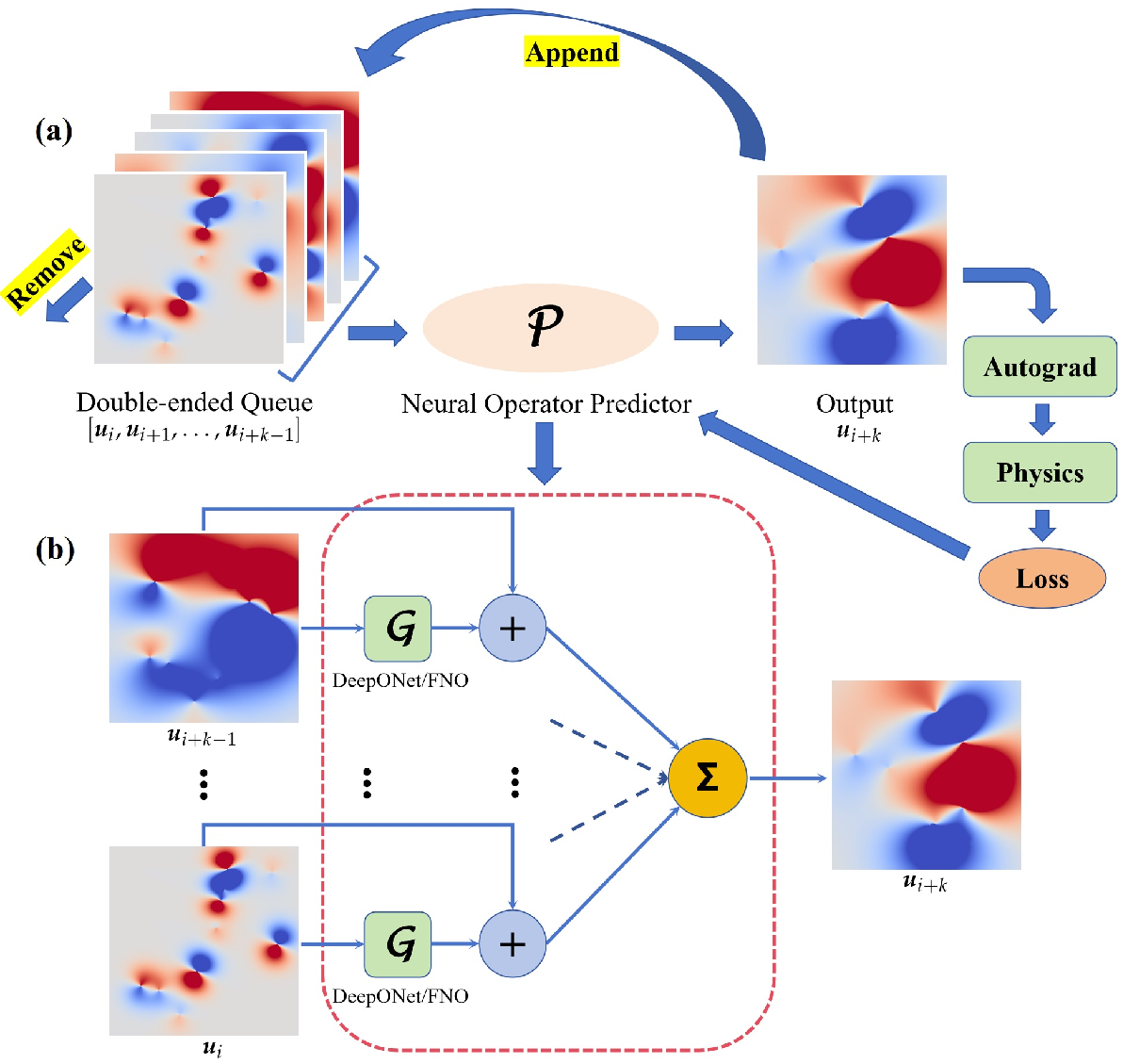}}}\hspace{-0.35in}
  \hspace{0in}
  \vspace{0.1in}
\caption{\small \rm The architectural overview of the PMNO. (a) Forward process and backpropagation. (b) Structure of the multi-step neural operator predictor.}
  \label{f-model}
\end{figure}

{\bf Forward process.}
Given that our prediction relies on past information, we reexpress $\bu_{i+k}$ accordingly in Eq.~(\ref{bash}) and implement the method using an explicit formulation, i.e. $b_k=0$ and $a_k\neq0$,
\begin{equation}\label{bash1}
  \bu_{i+k} = \sum_{j=0}^{k-1}\left(\alpha_j \bu_{i+j} + \beta_j\mathcal{N}[\bx,t_{i+j};\bu_{i+j}]\right),
\end{equation}
where $\alpha_j = -a_j/a_k$ and $\alpha_j=\Delta t b_j/a_k$.
Motivated by Eq.~(\ref{bash1}), we extend the neural operator framework by incorporating multiple past solution states into the input and introducing a residual connection, leading to the development of a multi-step neural operator, as shown in Fig.~\ref{f-model}(b),
\begin{equation}\label{MO}
  \mathcal{P}[\bu_{i}, \bu_{i+1}, \ldots, \bu_{i+k-1}]:= \sum_{j=0}^{k-1}\left[\lambda_j\bu_{i+j}(\bx) + \Delta t\delta_j\mathcal{G}(\bu_{i+j})(\bx)\right],
\end{equation}
where $\{\lambda_j\}_{j=0}^{k-1}$ and $\{\delta_j\}_{j=0}^{k-1}$ are the learnable weight parameters, $\Delta t$ refers to a fixed time step,  $\mathcal{G}$ represents the neural operator, which can be implemented using architectures such as DeepONet or FNO. Further details on the neural operator are provided in \ref{A-NO}.
Here $\lambda\bu + \Delta t\delta\mathcal{G}(\bu)$ can be regarded as a form of residual connection.
In fact, numerous studies \cite{resnet1,Polynet,fractalnet,multi-res} have shown that ResNet \cite{resnet}, one of the most influential deep learning architectures, closely resembles the forward Euler time integration scheme.
The multi-step neural operator advances this framework by integrating multi-step inputs, which improves its capacity for long-range prediction. And in terms of numerical error, linear multi-step methods are known to yield higher-order accuracy than the forward Euler scheme, reducing error accumulation over time.
Furthermore, according to the universal approximation theorem for neural operators \cite{deeponet,universal-deeponet,universal-fno,universal-fno1}, $\mathcal{G}$ can approximate any given continuous operator.
Consequently, with properly weight parameters, the prediction error can be effectively controlled at each stage.

The form of Eq.~(\ref{MO}) provides the foundation for designing a recurrent structure for continuous spatiotemporal prediction of the solution.
We adopt a Double-ended Queue, where at each forward step, both the head and tail are simultaneously remove and append the solution states as shown in Fig.~\ref{f-model}(a).
Meanwhile, the generated solution states allow for additional operations, which form the foundation for the subsequent backpropagation.

{\bf Backpropagation.}
By incorporating physical knowledge during backpropagation, the model is guided to learn the inherent physical principles governing the system.
Unlike the forward process, backpropagation allows access to intermediate outputs at each stage, making it straightforward to incorporate implicit multi-step methods into the training of the model. This integration enhances stability and reduces the dependence on large-scale training data.
This approach parallels the philosophy of backpropagation in machine learning, where intermediate outputs are exploited to continuously adjust the model parameters, driving predictions toward more accurate solutions.

For each stage output $\bu_{i+k}$ with $i=0, 1, \ldots$, $\mathcal{N}[\bu_{i+k}]$ can be computed either via automatic differentiation or numerical finite differences in neural operators \cite{PI-fno,PI-deeponet}. Therefore, we learn the network parameters by minimizing the following physics-guided loss function defined as
\begin{equation}\label{Loss}
  \mathcal{L}(\theta) = \frac{1}{L}\sum_{i=0}^{L-1}\mathcal{L}_i = \frac{1}{L}\sum_{i=0}^{L-1}\left\| \sum_{j=0}^{k}\left(a_j\bu_{i+j} - \Delta t b_j\mathcal{N}[\bu_{i+j}]\right) \right\|_2^2,
\end{equation}
where $L$ represents the number of recurrent iterations within the model, $\{a_j\}_{j=0}^k$ and $\{b_j\}_{j=0}^k$ are known fixed constants, determined by the linear multi-step method.
$\theta$ denotes the model parameters, which include the weights $\{\lambda_j\}_{j=0}^{k-1}$, $\{\delta_j\}_{j=0}^{k-1}$ and the parameters in neural operator $\mathcal{G}$.
And the norm $\|\cdot\|_2^2$ can be computed through a Monte Carlo approximation over sampled points.
$\{\bu_j\}_{j=0}^{k-1}$ are the only labeled data needed, and $\{\bu_j\}_{j=k}^{k+L-1}$  denote the model output at each stage
($\hat{\bu}_{i+k}:=\mathcal{P}[\hat{\bu}_{i}, \hat{\bu}_{i+1}, \ldots, \hat{\bu}_{i+k-1}]$), and for notational convenience, we use the same notation as the labeled data.
Among the various effective implicit methods, we primarily adopt the Backward Differentiation Formula (BDF) \cite{BDF} due to its favorable stability properties for training our neural operator model. Besides, it involves only the information of current-step function $\mathcal{N}[\bx, t_{i+k}; \bu_{i+k}]$ on the right-hand side, which facilitates the computation of the loss.
Under this setting, the loss function can be reformulated as,
\begin{equation}\label{Loss-bdf}
  \mathcal{L}(\theta) = \frac{1}{L}\sum_{i=0}^{L-1}\mathcal{L}_i = \frac{1}{L}\sum_{i=0}^{L-1}\left\| \sum_{j=0}^{k}a_j\bu_{i+j} - \Delta t b_k\mathcal{N}[\bu_{i+k}] \right\|_2^2.
\end{equation}
The detailed parameters about BDF can be found in the \ref{A-BDF}.

\subsection{Hard-constraint boundary conditions}
In the loss function, we do not impose any BC constraints. In fact, the BC can be strictly imposed in neural networks as hard-constraints.
Enforcing exact BCs in neural networks has been widely used \cite{bc1,ednn,bc2}. These techniques are also applicable to neural operators.

For the Dirichlet BC $\mathcal{B}[\bx,t;\bu(\bx,t)]=\bu(\bx,t)-g(\bx)=0$ at $\bx\in\partial\Omega$, we can construct the output of neural operator at each time step in the form of
$\hat{\bu}_{i+k}:=g(\bx)+\mathcal{P}[\bu_{i}, \bu_{i+1}, \ldots, \bu_{i+k-1}]p(\bx)$, where $p(\bx)=0$ at $\bx\in\partial\Omega$.
For the periodic BC with period $P$, FNO inherently enforces periodic boundary conditions on the output, making it a natural choice in such settings.
In the case of DeepONet, we can introduce the special feature expansion in trunk net in the form of $t_i(v(\bm y))$ with periodic function satisfying $v(\bm y+P)=v(\bm y)$.
For example, in the two-dimensional case, we can replace the trunk net input with Fourier basis functions, i.e.,
\begin{equation}\label{vxy}
  v(x,y)=\left(
           \begin{array}{ccc}
             \cos x\cos y, & \ldots, & \cos(M_x x)\cos(M_y y) \\
             \cos x\sin y, & \ldots, & \cos(M_x x)\sin(M_y y) \\
             \sin x\cos y, & \ldots, & \sin(M_x x)\cos(M_y y) \\
             \sin x\sin y, & \ldots, & \sin(M_x x)\sin(M_y y) \\
           \end{array}
         \right),
\end{equation}
and $M_x$, $M_y$ are some positive integers.

\subsection{Training and extrapolation}

In this work, the operator $\mathcal{G}$ is mainly instantiated using two architectures: DeepONet and FNO.
In the case of DeepONet, the input $\bm y$ of trunk net and the sensors $\{\bx_1, \bx_2, \ldots, \bx_m\}$ in branch net are the same for all $\bu$, where $N$ sample points $\{\bm y_i\}_{i=1}^N$ are randomly selected within the domain $\Omega$, and sensor locations are determined based on the type of the branch network. In fact, for DeepONet, the selection of sample points and sensors is more flexible. The sampling locations can be adaptively updated according to the output at each time step.
In contrast, due to architectural constraints, FNO requires both sampling points and function discretization to be defined on grid points.
We take $[\bu_{i}, \bu_{i+1}, \ldots, \bu_{i+k-1}, \bx]$ as the input data, where $\bu_{i+j} = (\bu_{i+j}(\bx_1), \bu_{i+j}(\bx_2), \ldots, \bu_{i+j}(\bx_N))^T$ and $\bx = (\bx_1,\bx_2,\ldots, \bx_N)^T$ corresponds to uniformly grid points.

Training the loss defined in Eq.~(\ref{Loss-bdf}) is difficult for large $L$.
In many previous works \cite{pdenet,RD-nmi}, a multi-stage training strategy was adopted, where the model is first pre-trained using a small number of time steps, and then gradually fine-tuned on longer horizons. The primary challenge in training, in fact, arises from the fact that the optimization process often violates the temporal causality such that the loss $\mathcal{L}_i$ will be minimized even if the predictions at $t_{i+k}$ and previous times are inaccurate. This issue is particularly critical for sequence learning models, as it can lead to the accumulation of errors and ultimately cause the model to learn incorrect dynamics. Therefore, we adopt a causal training strategy \cite{causal pinn} by defining a weighted residual loss as
\begin{equation}\label{causal loss}
  \mathcal{L}(\theta) = \frac{1}{L}\sum_{i=0}^{L-1}\omega_i\mathcal{L}_i,
\end{equation}
with
\begin{equation}\label{wi}
  \left(
  \begin{array}{c}
    \omega_0 \\
    \omega_1 \\
    \omega_2 \\
    \vdots \\
    \omega_{L-1} \\
  \end{array}
\right) = \exp\left[-\epsilon \left(
\begin{array}{ccccc}
0 & 0 & 0 & \cdots & 0 \\
1 & 0 & 0 & \cdots & 0 \\
1 & 1 & 0 & \cdots & 0 \\
\vdots & \vdots & \vdots & \ddots & \vdots \\
1 & 1 & 1 & \cdots & 0
\end{array}
\right)
\left(
  \begin{array}{c}
    \mathcal{L}_0 \\
    \mathcal{L}_1 \\
    \mathcal{L}_2 \\
    \vdots \\
    \mathcal{L}_{L-1} \\
  \end{array}
\right)
  \right],
\end{equation}
where $\epsilon$ denotes a given parameter that governs the steepness of the weights $\omega_i$ during the optimization process.
Observe that $\omega_i$ shrinks exponentially with the accumulated residual errors from earlier time steps, which means that the loss term $\mathcal{L}_i$ can only be effectively minimized once the residuals $\{\mathcal{L}_j\}_{j=0}^{i-1}$ from earlier steps have sufficiently decreased.
Causal training prevents the network from converging to local minima, while also eliminating the need for multi-stage training strategies, thereby accelerating the overall training process.
And the computational cost is negligible since the weights $\omega_i$ are computed via the direct accumulation of loss values $\{\mathcal{L}_j\}_{j=0}^{i-1}$ from previous time steps
whose values are already stored in the computational graph during training.

Owing to the neural operator framework, our model supports extrapolation across arbitrary spatial resolutions.
When adopting the DeepONet framework for $\mathcal{G}$, high-resolution spatial sampling is required solely for the trunk network, while the sensor positions in the branch network remain unchanged. Subsequently, a recurrent structure enables continuous solution prediction at this resolution.
In the case of FNO, the initial input $[\bu_{0}, \bu_{1}, \ldots, \bu_{k-1}]$ need to be sampled on a high-resolution grid and used as the model input with grid points.
Similarly, the recurrent structure enables prediction of the solution over all future time steps.
The framework of PMNO predictor, including model construction, training, and extrapolation, is summarized in Algorithm \ref{alg1}.

\begin{algorithm}[ht]
    \caption{The framework of PMNO predictor.}\label{alg1}
    \begin{algorithmic}
      \v  \REQUIRE  The order $k$ of the time integrator; time interval $\Delta t$; the number of recurrent iterations $L$; the fixed parameters determined by $k$-step BDF $\{a_j\}_{j=0}^k$ and $b_k$; the initial input $[\bu_{0}, \bu_{1}, \ldots, \bu_{k-1}]$ and sample points $\{\bx_i\}_{i=1}^N$.
       \v \ENSURE Output $\hat{\bu}_{i},i=0,1,\ldots, M$ on a high-resolution grid.
        \STATE \textbf{Model}: Construct the multi-step neural operator predictor $\mathcal{P}$ with initialized parameter
        \begin{equation*}
          \mathcal{P}: = \sum_{j=0}^{k-1}\left[\lambda_j\bu_{i+j}(\bx) + \Delta t\delta_j\mathcal{G}(\bu_{i+j})(\bx)\right].
        \end{equation*}

        \v \STATE \textbf{Forward process}:
             Initialize $U = [\bu_{0}, \bu_{1}, \ldots, \bu_{k-1}]$.
        \FOR{$i=0:L$}
            \STATE Predict the next time step:
            \begin{equation*}
              \bu_{i+k} = \mathcal{P}(U).
            \end{equation*}
            \STATE Double-ended Queue:
            \begin{equation*}
              U = [\bu_{i+1}, \bu_{i+2}, \ldots, \bu_{i+k}].
            \end{equation*}
    \ENDFOR

    \v \STATE \textbf{Backpropagation}:
         Compute the loss function at each time step $\mathcal{L}_i$ using the forward process with $\{a_j\}_{j=0}^k$ and $b_k$:
        \begin{equation*}
          \mathcal{L}_i=\left\| \sum_{j=0}^{k}a_j\bu_{i+j} - \Delta t b_k\mathcal{N}[\bu_{i+k}] \right\|_2^2=\frac{1}{N}\sum_{l=1}^{N}\left(\sum_{j=0}^{k}a_j\bu_{i+j}(\bx_l) - \Delta t b_k\mathcal{N}[\bu_{i+k}](\bx_l)\right)^2.
        \end{equation*}
        \STATE Compute the weighted residual loss:
        \begin{equation*}
          \mathcal{L}(\theta) = \frac{1}{L}\sum_{i=0}^{L-1}\omega_i\mathcal{L}_i.
        \end{equation*}
        \STATE Perform optimization over $\{\lambda_j\}_{j=0}^{k-1}$, $\{\delta_j\}_{j=0}^{k-1}$ and the internal parameters of the neural operator $\mathcal{G}$.

        \v \STATE \textbf{Extrapolation}:
          Sample the initial input $[\bu_{0}, \bu_{1}, \ldots, \bu_{k-1}]$ on a high-resolution grid.
        \STATE Output $\bu_{i}$, $i=0,1,\ldots, M$ on the high-resolution grid through the forward process.

    \end{algorithmic}
\end{algorithm}

\begin{table}[!ht]
\vspace{0.05in}
\begin{center}\small
\caption{\small \rm The valid predicted times $T_v$ for different physical systems through different methods, where ``/" represents that the method either failed to converge during training or that $\varepsilon(t)\geq \tau$ holds universally.}\label{table-results}
\vspace{0.05in}
\begin{tabular}{ c|cccccc}
\Xhline{1pt}
\rule{0pt}{10pt}         \diagbox{Models}{Systems}            &  Advection   &  $\lambda$-$\omega$ RD  &  Heat   & NLS & GS  \\[0.5ex]
\Xhline{1pt}

\rule{0pt}{10pt}  DeepONet-MLP       &   /                & /                & 3.03           & /              & / \\[0.5ex]
\hline

\rule{0pt}{10pt}  DeepONet-CNN       &   /                & 8.35             & /              & /              & 46 \\[0.5ex]
\hline

\rule{0pt}{10pt}   FNO               &   2.22             & 3.25             & 2.16           & 9.6            & 76 \\[0.5ex]
\hline

\rule{0pt}{10pt}   ConvLSTM          &   3.74             & 8.25             & 1.32           & /              & 47.2 \\[0.5ex]
\hline

\rule{0pt}{10pt}   PIdeepONet-MLP    &   /                & /                & \textbf{6.15}  & /              & / \\[0.5ex]
\hline

\rule{0pt}{10pt}   PIdeepONet-CNN    &   /                & \textbf{24}      & /              & 10.95          & 65 \\[0.5ex]
\hline

\rule{0pt}{10pt}   PIFNO             &   \textbf{4.64}    & 3.9              & 3.33           & \textbf{30.9}  & \textbf{82} \\[0.5ex]
\hline

\Xhline{1pt}
\end{tabular}
\end{center}
\end{table}

\section{Numerical experiments}

In this section, we demonstrate the superior predictive capabilities of PMNO predictor by experimenting with a wide range of physical systems, including the 2D advection equation, $\lambda$-$\omega$ reaction-diffusion (RD) equation, the 2D heat equation with an irregular domain, the 3D nonlinear Schr\"{o}dinger (NLS) equation and the Gray–Scott (GS) model, in comparison to the general operator based methods and fixed-resolution approaches.
To ensure fairness, we employ the multi-step neural operator structure for the general operator-based prediction methods in the following examples, highlighting the key role of implicit physics guidance in backpropagation. Additionally, we consider the ConvLSTM method \cite{convlstm} for comparison. On one hand, its recurrent convolutional structure shares similarities with our network, and on the other hand, it outperforms other fixed-resolution methods, such as ResNet, PDE-net, and others \cite{resnet,pdenet,transformer-nn}.
All the data and codes are available in GitHub at https://github.com/JinSong99/PMNO.

The network structures, hyperparameters and model parameters involved in the different examples are summarized in the tables in \ref{A-parameters}.
The Adam optimizer \cite{adam} is adopted to optimize the weighted residual loss as in Eq.~(\ref{causal loss}).
And the learning rate is initialized to $\eta_0$ = 1e-3 and decays by 0.99 every 200 iterations.
Unless otherwise specified, we set $k=5$ in this section to evaluate our method across all physical systems.
The impact of the step size $k$ on the performance of the predictor will be discussed in detail in the next section.
Since our primary focus is on prediction horizon, we define the following instantaneous relative root-mean-square error (RMSE) $\varepsilon(t)$ to test the validity and advantages of our method,
\begin{equation}\label{rmse}
  \varepsilon(t_i):=\frac{\left\|\bu(t_i) -\bu^*(t_i)\right\|_{2}}{\left\|\bu(t_i)\right\|_{2}},
\end{equation}
where $\bu^*(t_i)$ is the reference solution at $t_i=i\Delta t$ including training and extrapolation phases.
Furthermore, we measure reliable prediction horizon using the associated valid time, denoted as $T_v$. This is defined as the first time step at which the error $\varepsilon(t)$ exceeds a certain threshold $\tau$, that is
\begin{equation}\label{T-valid}
  T_v=\inf_t\{t|\varepsilon(t)\geq\tau\}.
\end{equation}
We set $\tau=0.1$ and  evaluate the predictive performance of the ConLSTM method, DeepONet and FNO-based methods, and our proposed PMNO approach. For DeepONet, both the vanilla version (hereinafter referred to as DeepONet-MLP) and the DeepONet-CNN variant \cite{deeponet-fair,deeponet-cnn} are considered. The valid predicted times of all models are summarized in Table \ref{table-results}. As is shown, the physics-guided implicit BDF-driven model exhibits superior predictive performance compared to data-driven approaches.
The implementation details of each example are presented in the following.


\subsection{\it 2D advection equation} \label{ex1}

\begin{figure}[!t]
    \centering
      {\scalebox{0.77}[0.77]{\includegraphics{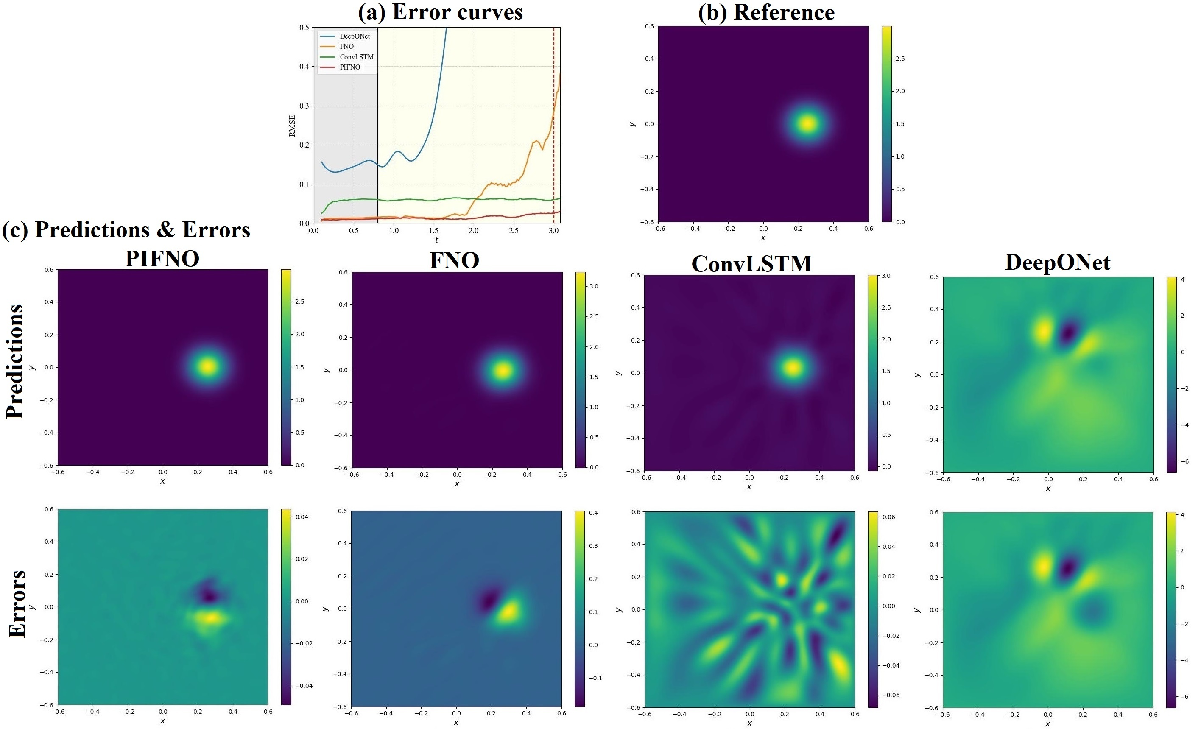}}}
  \hspace{0in}
  \vspace{-0.1in}
\caption{\small \rm  2D advection equation: (a) Error propagation curves, where the gray region on the left represents the training domain, while the yellow one on the right indicates the extrapolation domain. (b) The reference solution at $t=3$. (c) The predictions and corresponding errors (error $=u-u^*$) from each predictor at $t=3$.}
  \label{f-2Dadvection}
\end{figure}

The first example we consider is the classic 2D advection equation, governed by the following linear PDE in the form of \cite{ad}
\begin{equation}\label{2Dadvection}
 \frac{\partial u(\bx,t)}{\partial t}=(\mathbf{a}(t)\cdot \nabla)u(\bx,t),
\end{equation}
where $\bx=(x,y)\in [-l,l]\times[-l,l]$, $\mathbf{a}(t)=(a_1(t),a_2(t))$ is the flow velocity, and it is only dependent on time.
Especially, we specify the true solution as
\begin{equation}\label{ad-solu}
  u(\bx,t)=\exp\left[-A[(x-v_1(t))^2+(y-v_2(t))^2]\right],
\end{equation}
where $A>0$. We can verify that when $a_i(t)=-\dot{v_i}(t)$, the $u(\bx,t)$ above is the solution of Eq.~(\ref{2Dadvection}).
It should be noted that when $A$ is large, the solution exhibits local behavior only near the point $(v_1(t),v_2(t))$ and decreases rapidly away from this point.
Hence, the solution can be assumed to vanish at the boundary, satisfying periodic boundary conditions.
Herein, we choose $l=0.6$, $A=100$, and $(v_1(t), v_2(t))=(0.25\cos(2\pi t), 0.25\sin(2\pi t))$ for performance comparison.

With the time interval $\Delta t=0.02$ and $L=36$, the training time duration is $[0,0.8]$, that is $(L+k-1)\Delta t$.
The solution error propagation with time $\varepsilon(t)$ for each method is displayed in Fig.~\ref{f-2Dadvection}(a).
Among them, we observe that the DeepONet-based models, including DeepONet-MLP and DeepONet-CNN, exhibit relatively poor predictive performance, which may be attributed to the localized behavior of the solution. To better illustrate the extrapolation capability of PMNO method, the snapshots at $t=3$ of predictions and errors by each method are shown in Fig.~\ref{f-2Dadvection}(c) in comparison with the ground truth reference (Fig.~\ref{f-2Dadvection}(b)). It can be seen that the physics-guided neural operator model exhibits superior extrapolative predictive capability with the relative RMSE below 3\%. Although PIFNO-based operator model performs similarly to FNO during the training phase, it exhibits superior extrapolation capability, enabling accurate prediction of dynamical behavior over longer time horizons. Specifically, the valid predicted time is up to twice as long as that of purely data-driven models (see Table \ref{table-results}).

Notably, while our operator model is trained at a resolution of $128\times128$ (as a mesh-free method, DeepONet-MLP is implemented with $N=3000$ sample points), extrapolation is performed at a higher resolution of $256\times256$. Even under this situation, the PMNO method exhibits superior predictive performance compared to fixed-resolution baselines such as ConvLSTM method.
This is attributed to the discretization-invariant property of the neural operator based models \cite{universal-fno}.

\subsection{\it $\lambda$-$\omega$ reaction-diffusion equations}

\begin{figure}[!t]
    \centering
  {\scalebox{0.5}[0.5]{\includegraphics{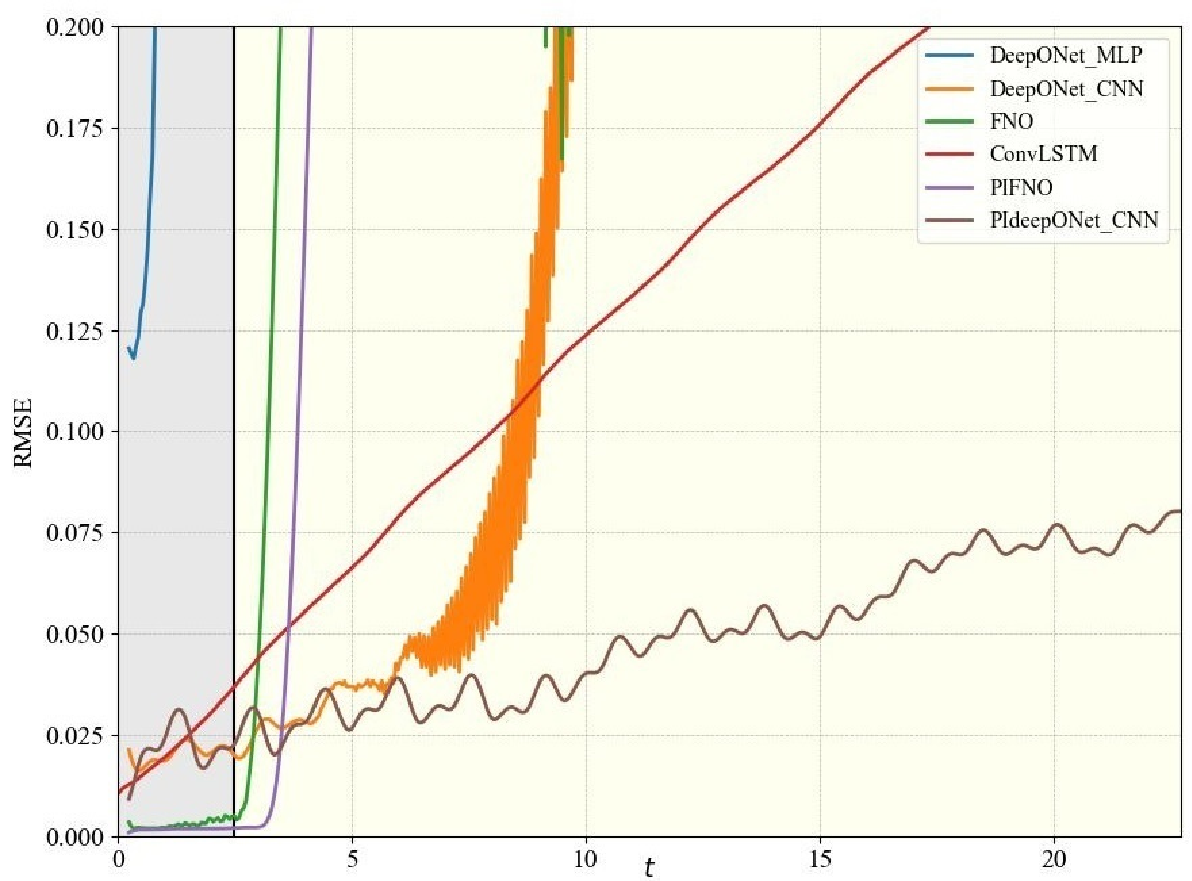}}}
  \vspace{-0.1in}
\caption{\small \rm $\lambda$-$\omega$ RD equations: Error propagation curves, where the gray region on the left represents the
training domain, while the yellow one on the right indicates the extrapolation domain.}
  \label{f-2DRD}
\end{figure}

\begin{figure}[!t]
    \centering
  {\scalebox{0.77}[0.77]{\includegraphics{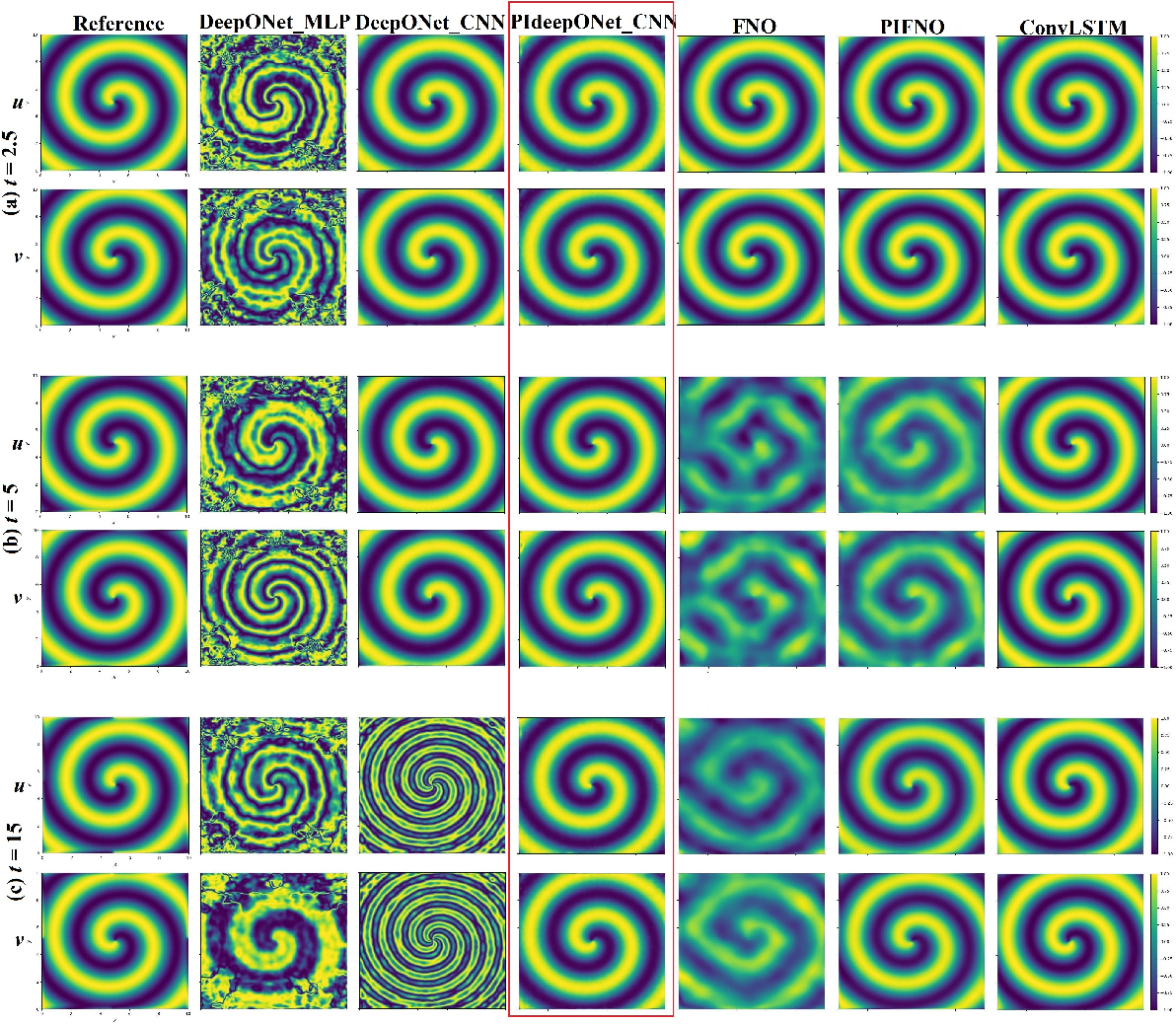}}}
  \vspace{-0.1in}
\caption{\small \rm $\lambda$-$\omega$ RD equations: The reference solutions and the predictions from each predictor at different times $t = 2.5$, $t=5$ and $t=15$.}
  \label{f-2DRDuv}
\end{figure}

The second example involves the reaction-diffusion (RD) system in a 2D domain, which is well known for wide applications in population dynamics, chemical reactions and so on \cite{RD1,RD2,RD3}.
Specifically, we consider the $\lambda$-$\omega$ RD equations with the formation of spiral wave patterns that can be described by the following coupled nonlinear PDEs \cite{RD3}:
\begin{equation}\label{2DRD}
  \begin{split}
      & u_t=D\Delta u + (\lambda u - \omega v)-(u^2+v^2)u, \\
      & v_t=D\Delta v + (\omega u + \lambda v)-(u^2+v^2)v,
  \end{split}
\end{equation}
where $\bu=[u,v]^T$, $D$ is the diffusion coefficient, $\lambda$ and $\omega$ are fixed parameters.
Herein, we choose $D=0.001$, $\lambda=0.98$ and $\omega=1$ to evolve the spiral wave solution on domain $\Omega=[0,10]\times[0,10]$ by a spectral method \cite{spectral} as a reference.
Considering the periodic patterns of solutions, we adopt the sine function as the activation function in the DeepONet-based model.
And DeepONet-MLP is also extended to multiple output functions, where the output of branch net is split to two groups and the trunk net is shared. The specific data shape and network architectures can be found in tables in \ref{A-parameters}.

We train the model in time duration of $[0, 2.5]$ with $\Delta t=0.05$. The error propagation with time $\varepsilon(t)$ for each method is shown in Fig.~\ref{f-2DRD}.
The period boundary conditions are incorporated into the DeepONet-based model by Eq.~(\ref{vxy}) with $M_x=M_y=7$. It can be seen that DeepONet-CNN exhibit a stronger ability to capture global features in comparison to DeepONet-MLP, which facilitates model convergence during training.
In contrast, although the FNO-based model achieves low error during the training phase, its extrapolation capability is notably poor.
Notably, PIdeepONet-CNN is able to capture the intrinsic dynamics of the system. In contrast to the purely data-driven DeepONet, it achieves substantially longer extrapolation, with the valid predicted horizon $T_v=24$, nearly ten times longer than the training time duration.
Furthermore, the reference solutions and the predictions from each predictor are shown in Fig.~\ref{f-2DRDuv} at different times $t = 2.5$, $t=5$ and $t=15$, where the extrapolation is performed at the resolution of $256\times256$.
It can be observed that both the training and extrapolation results obtained by PIDeepONet-CNN exhibit excellent agreement with the reference solution, which is highlighted with a red box in Fig.~\ref{f-2DRDuv}.

\begin{figure}[!ht]
    \centering
  {\scalebox{0.77}[0.77]{\includegraphics{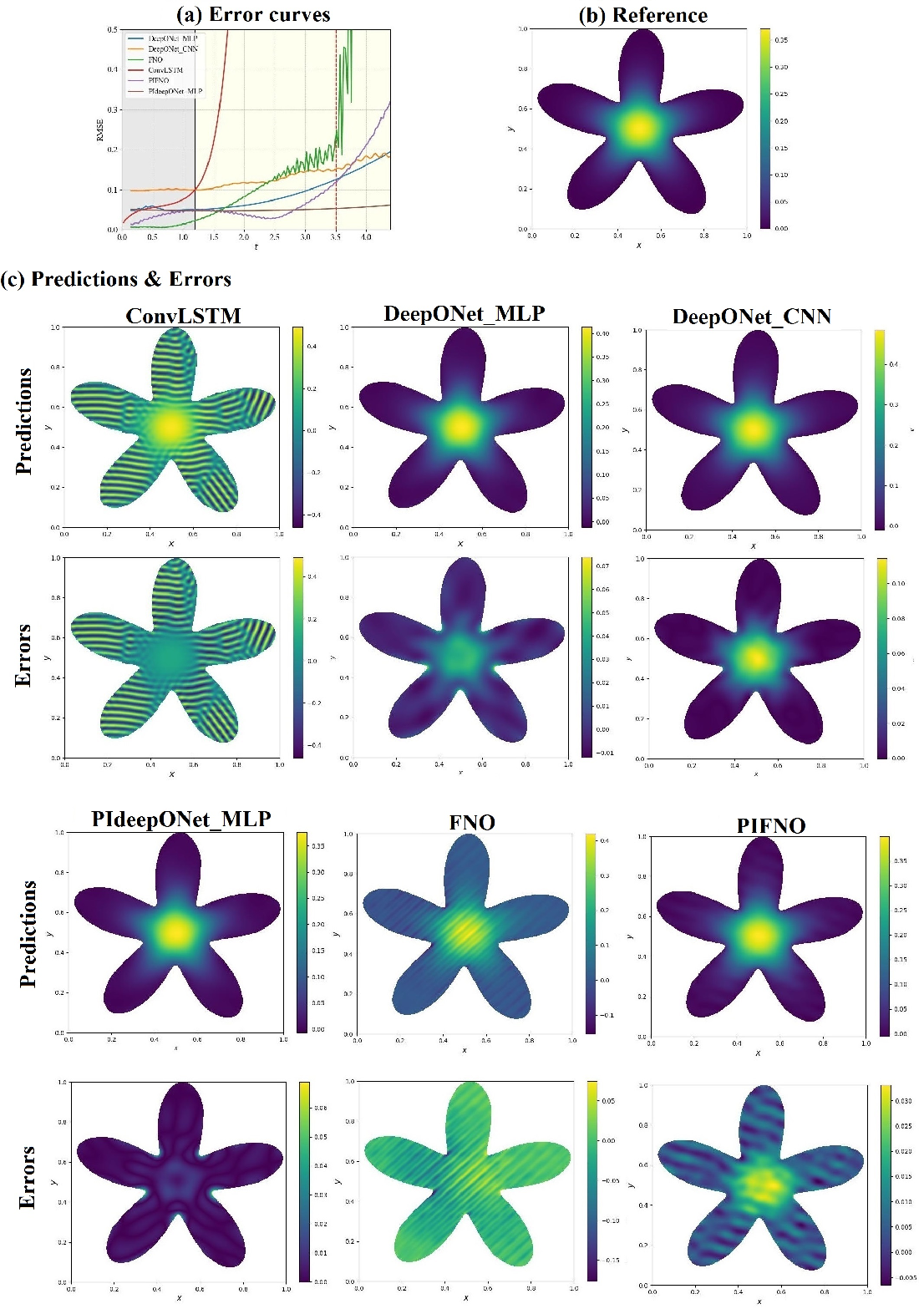}}}
  \vspace{-0.1in}
\caption{\small \rm 2D heat equation with an irregular domain: (a) Error propagation curves, where the gray region on the left represents the training domain, while the yellow one on the right indicates the extrapolation domain. (b) The reference solution at $t=3.5$. (c) The predictions and corresponding errors (error $=u-u^*$) from each predictor at $t=3.5$.  }
  \label{f-2Dheat}
\end{figure}

\subsection{\it 2D heat equation with an irregular domain}

Then the example we consider is the 2D heat equation in an irregular domain, which can be governed by the following equation
\begin{equation}\label{2Dheat}
  u_t=\alpha \nabla^2 u, \quad (x,y)\in \Omega
\end{equation}
where $\alpha$ is thermal diffusivity, and $\Omega$ is taken as a five-pointed star-shaped domain, which can be explicitly described in polar coordinates as follows
\begin{equation}\label{domain}
  \Omega=\left\{(r,\theta)| 0\leq r\leq R\frac{1+0.5\sin(5\theta)}{1+0.5|\sin(2.5\theta)|} \right\},
\end{equation}
where $x=x_0+r\cos\theta$ and $y=y_0+r\sin\theta$. We set $x_0=y_0=0.5$, $R=0.38$, $\alpha=0.001$, and consider the Dirichlet boundary condition, i.e., $u=0$ for $(x,y)\in\partial \Omega$.

Considering that convolution-based models including DeepONet-CNN and FNO require grid-structured data, we design a masking module to automatically set the values outside the domain to zero.
Nevertheless, gridding the domain necessitates relatively high-resolution data to properly resolve boundary effects. In this example, the model is trained on data with a spatial resolution of $256\times256$. The time interval is set to $\Delta t = 0.03$, and the training time duration is $[0, 1.2]$.

The error propagation with time $\varepsilon(t)$ for each method is shown in Fig.~\ref{f-2Dheat}(a). And the predictions and corresponding errors from each predictor at $t=3.5$ are exhibited in Fig.~\ref{f-2Dheat}(c), where the extrapolation is performed at the resolution of $1024\times1024$, except for the fixed-resolution ConvLSTM method.
It can be observed that for convolution-based methods, prediction accuracy degrades as the prediction horizon increases. The increased error can be attributed to the difficulty grid-based approaches face in accurately resolving complex boundary geometries.
And the relatively large training error observed in both DeepONet-MLP and PIDeepONet-MLP may be attributed to the Monte Carlo approximation used during training.
It is worth noting that the PIDeepONet-MLP model, with only $N=5000$ sampling points, is able to accurately capture the long-term dynamics of the heat equation. This can be attributed to the multi-step neural operator architecture and the BDF-guided training strategy.

\subsection{\it 3D complex-valued wave dynamics}

\begin{figure}[!t]
    \centering
  {\scalebox{0.7}[0.7]{\includegraphics{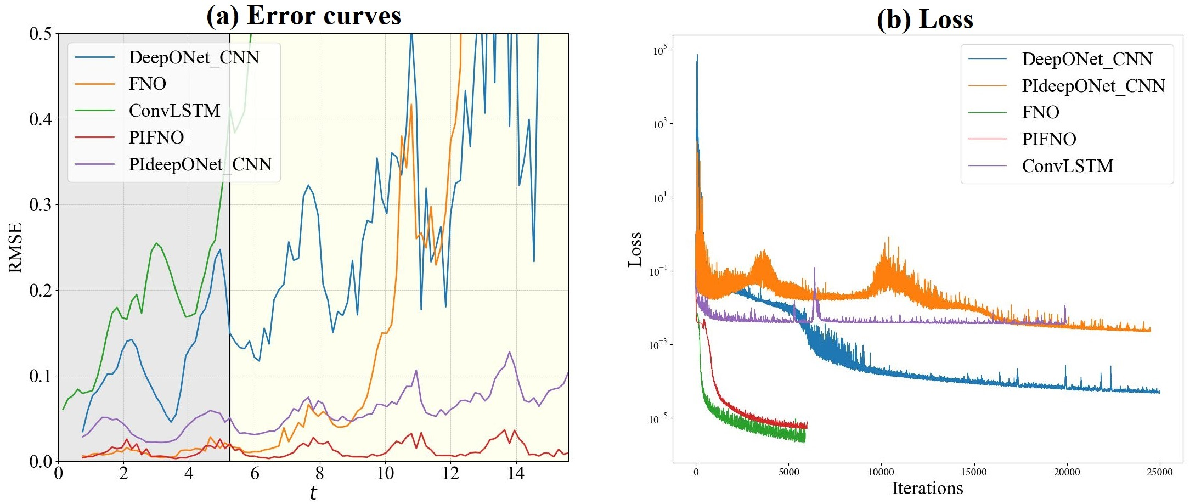}}}
  \vspace{-0.1in}
\caption{\small \rm 3D complex-valued wave dynamics: (a) Error propagation curves, where the gray region on
the left represents the training domain, while the yellow one on the right indicates the extrapolation domain. (b) The corresponding loss curves.}
  \label{f-3dnls}
\end{figure}

\begin{figure}[!t]
    \centering
  {\scalebox{0.8}[0.8]{\includegraphics{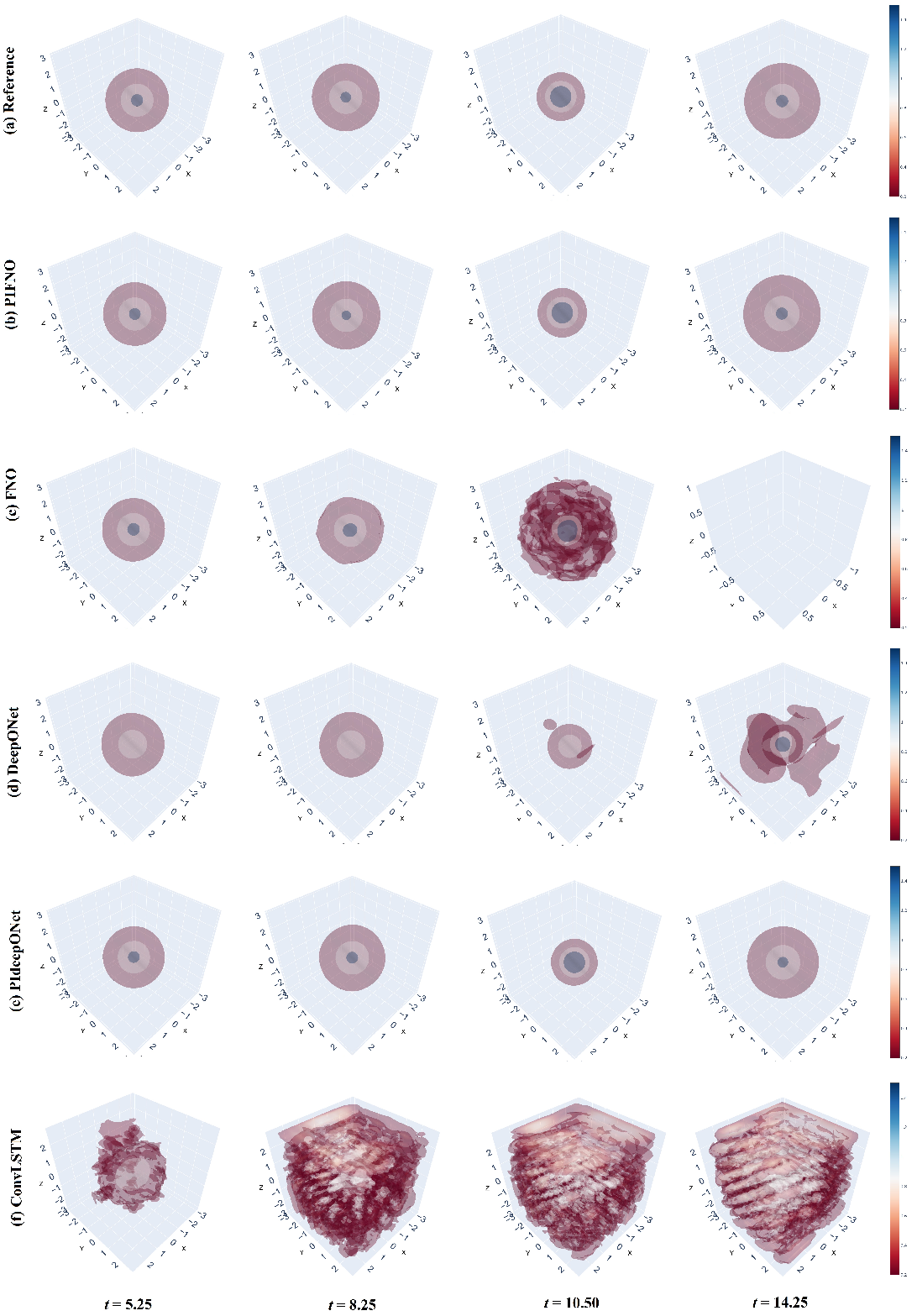}}}
  \vspace{-0.1in}
\caption{\small \rm 3D complex-valued wave dynamics: Isosurfaces of $|U|$ at values 0.2, 0.85, 1.5 for predictions and references from each predictor at different times $t=5.25$, $t=8.25$, $t=10.5$ and $t=14.25$. }
  \label{f-3dprediction}
\end{figure}

\begin{figure}[!t]
    \centering
  {\scalebox{0.77}[0.77]{\includegraphics{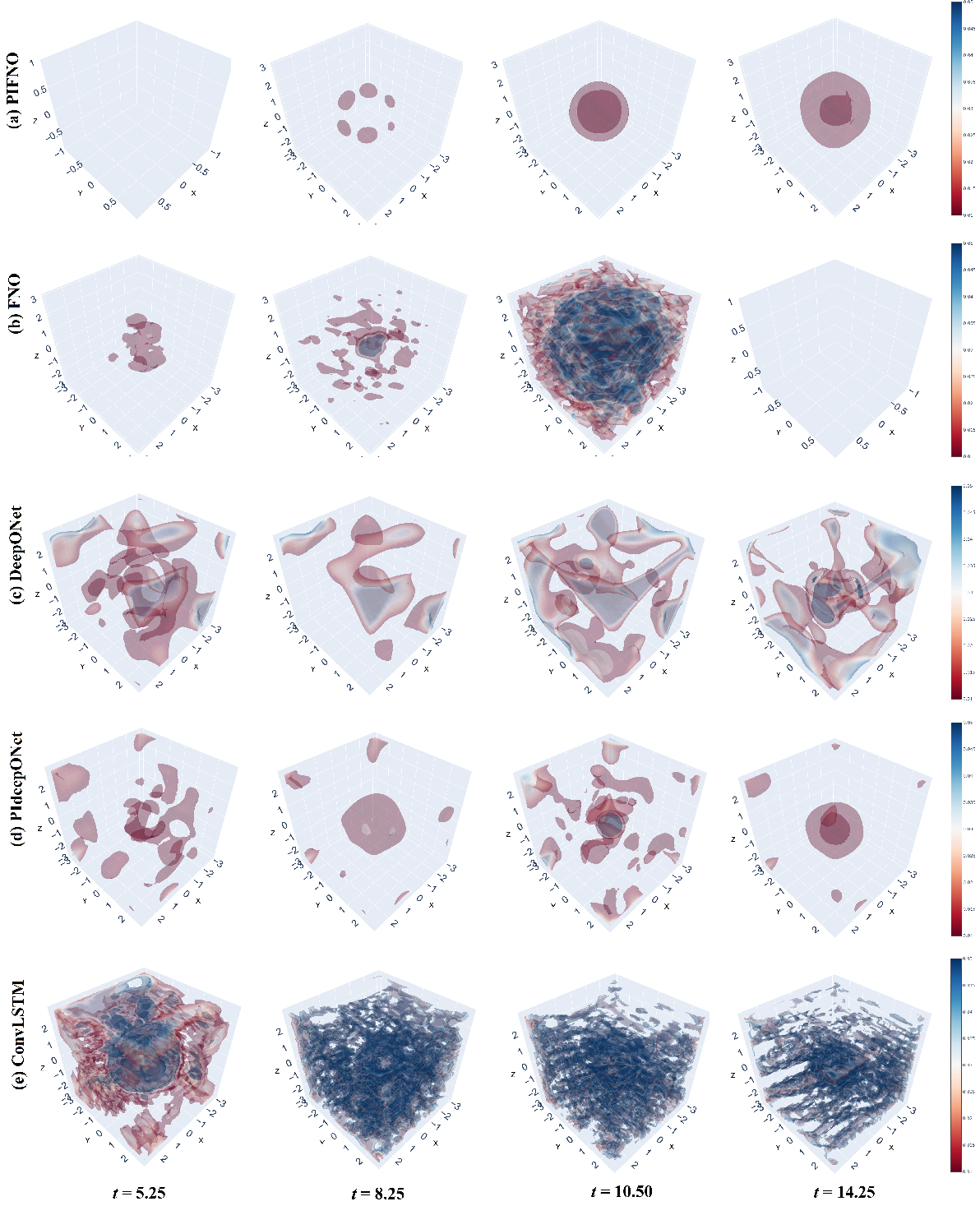}}}
  \vspace{-0.1in}
\caption{\small \rm 3D complex-valued wave dynamics: Isosurfaces of $|U-U^*|$ at values 0.01, 0.03, 0.05 for corresponding error against reference solution from each predictor at different times $t=5.25$, $t=8.25$, $t=10.5$ and $t=14.25$. }
  \label{f-3derror}
\end{figure}

In the next example, we consider the prediction of 3D complex-valued nonlinear wave dynamics, provided by an optical Kerr cavity
with diffraction and anomalous dispersion, with the addition of an attractive 3D parabolic potential \cite{3Dbreather}
\begin{equation}\label{3Dnls}
  U_t=i\Delta U-i(x^2+y^2+z^2)U+i|U|^2U-(1+i\delta)U+P,
\end{equation}
where $U(x, y, z,t)=u(x,y,z,t)+iv(x,y,z,t)$ represents the 3D complex-valued field amplitude, $z$ denotes the ``fast time" of the
field evolution within each round-trip, and $t$ corresponds to the ``slow time" that normalizes the round-trip
time of the cavity \cite{3Dbreather}, $\Delta=\partial_x^2+\partial_y^2+\partial_z^2$, where $\partial_z^2$ represents the group velocity dispersion.
Besides, $\delta$ denotes the detuning between the driving laser and the closest cavity resonance, $P$ signifies the pump amplitude, and $x^2+y^2+z^2$ describes the parabolic potential.
We take $\delta=-2$, $P = 0.75$ and set the continuous-wave $U=0.1e^{-(x^2+y^2)/a^2}$ with $a^2=2/\ln 2$ as the input. Over time, the continuous wave transitions into a stable breather state.

Considering that the solution is a complex-valued function, we transform Eq.~(\ref{3Dnls}) into a coupled system by setting $U=u+iv$,
\begin{equation}\label{3Dnls-coupled}
  \begin{split}
    u_t & =-\Delta v+(x^2+y^2+z^2)v-(u^2+v^2)v-u+\delta v +P, \\
    v_t & = \Delta u-(x^2+y^2+z^2)u+(u^2+v^2)u-v-\delta u.
  \end{split}
\end{equation}
With the time interval $\Delta t = 0.15$ and $L=31$, the training time duration is $[0, 5.25]$.
The error propagation with time $\varepsilon(t)$ for each method is shown in Fig.~\ref{f-3dnls}(a).
It can be seen that the physics guided neural operator methods exhibit superior predictive capability compared to data-driven models.
The effective prediction horizon of PIFNO reaches 30.9, which significantly surpasses that of the FNO model (see Table \ref{table-results}).
Moreover, we observe that for high-dimensional systems, FNO-based architectures tend to outperform those based on DeepONet.
In terms of training efficiency, FNO-based models also converge faster. As shown in Fig.~\ref{f-3dnls}(b), the FNO-based models reach convergence within only 6500 training steps.

We plot the prediction results of different methods, as well as the reference solutions at different times in Fig.~\ref{f-3dprediction}.
It can be seen that both PIDeepONet and PIFNO models effectively capture the characteristics of breathers, exhibiting clear periodic behavior.
Meanwhile, for better visualization, the isosurfaces of the corresponding error against reference solution from each predictor at different times are displayed in Fig.~\ref{f-3derror}.
While the error profile for PIFNO itself displays a breather-like oscillatory behavior, the relative error consistently stays under 4\%, with the absolute error remaining below 0.05, throughout the prediction horizon.
It should be noted that our operator model is trained at a resolution of $32\times 32\times32$ (for DeepONet-based models, details of the training data are provided in Table \ref{table-deeponetcnn} of the \ref{A-parameters}), extrapolation is performed at a higher resolution of $64\times 64\times64$.

\subsection{\it 3D Gray–Scott model} \label{ex5}

\begin{figure}[!t]
    \centering
  {\scalebox{0.7}[0.7]{\includegraphics{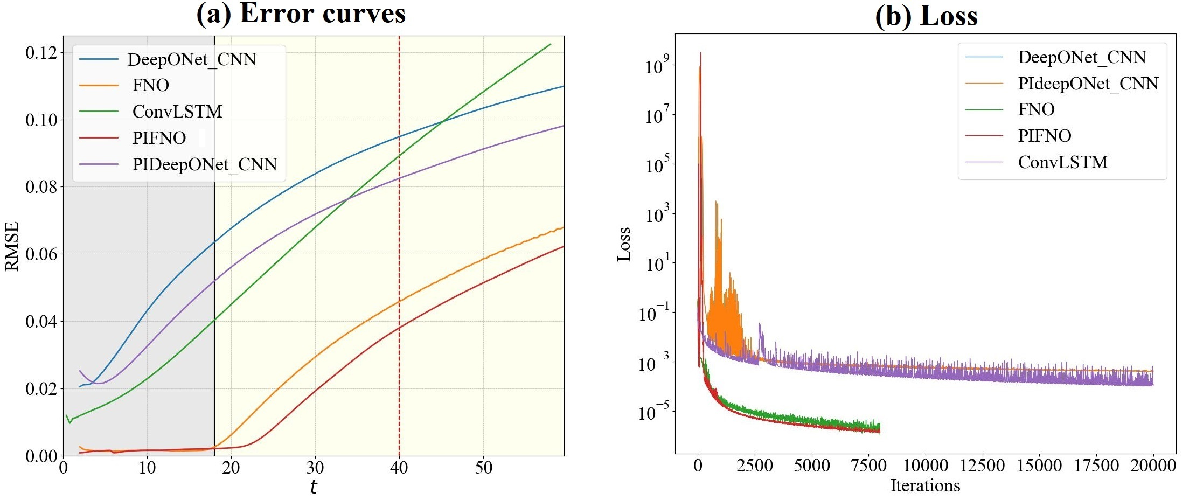}}}
  \vspace{-0.1in}
\caption{\small \rm 3D Gray–Scott model: (a) Error propagation curves, where the gray region on
the left represents the training domain, while the yellow one on the right indicates the extrapolation domain. (b) The corresponding loss curves. }
  \label{f-3DGS}
\end{figure}

\begin{figure}[!t]
    \centering
  {\scalebox{0.77}[0.77]{\includegraphics{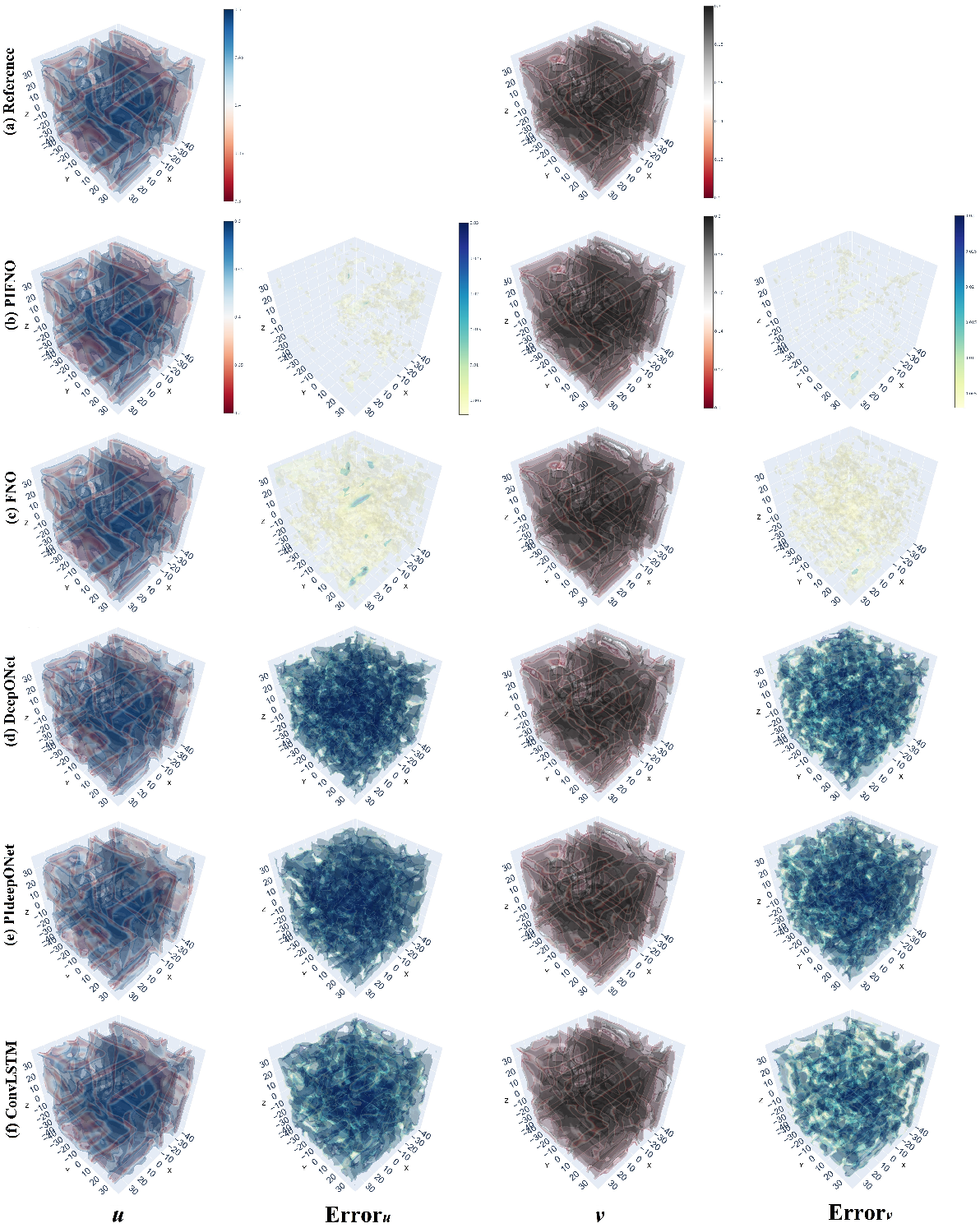}}}
  \vspace{-0.1in}
\caption{\small \rm 3D Gray–Scott model: Isosurfaces of $u$ at values 0.3, 0.4, 0.5 and $v$ at values 0.1, 0.15, 0.2 for each predictor and the corresponding error at values 0.002, 0.016, 0.03 against
reference solution at $t=40$. }
\label{f-3DGSuv}
\end{figure}

In the last example, we consider a more complex 3D reaction–diffusion model, Gray-Scott (GS) model in the form of
\begin{equation}\label{GS}
  \begin{split}
    u_t &= D_u \Delta u - auv^2 + F(1-u), \\
    v_t &= D_v \Delta v + auv^2 - (F+\kappa)v,
  \end{split}
\end{equation}
where $a$ is a scaling constant, $\kappa$ and $F$ represent the kill and feed rate respectively, $D_u$ and $D_v$ are diffusion coefficients.
Compared with the previous 2D $\lambda$-$\omega$ RD system, 3D GS model is capable of generating complex patterns such as spots, stripes, and spirals \cite{RD1,RD2,RD3}.
We set the model parameters with $D_u=4$, $D_v=2$, $a=20$, $F=0.5$ and $\kappa=1.1$, and consider a periodic boundary condition with $\Omega=[-40,40]^3$.

With the time interval set to $\Delta t = 0.4$ and $L=41$, the training time duration is $[0, 18]$. The error propagation with time $\varepsilon(t)$ for each method is exhibited in Fig.~\ref{f-3DGS}(a), which shows that physics-guided neural operator methods exhibits superior predictive capability compared to data-driven models.
Consistently, the FNO-based architecture demonstrates superior performance over its DeepONet counterpart.
Notably, it attains a relative training error as low as 2\%. More importantly, this advantage persists during the extrapolation phase, where the FNO-based model maintains significantly lower errors over longer prediction horizons with effective prediction time $T_v=82$.
It should be noted that a relatively large time interval is adopted, yet the PMNO model is still able to achieve accurate predictions, which can be attributed to our multi-step neural operator architecture.
With regard to training efficiency, we also observe that the FNO-based model is easier to train compared to the DeepONet-based model.
As depicted in Fig.~\ref{f-3DGS}(b), the loss for the DeepONet-based model begins at a large value and necessitates 20000 iterations to reach convergence, whereas the FNO-based model achieves convergence in only 8000 steps.

For better visualization, we also plot the prediction results and the corresponding error against reference solution at $t = 40$ in Fig.~\ref{f-3DGSuv}.
In terms of the absolute error, the PIFNO model exhibits excellent agreement with the ground truth across both training and extrapolation phases with the absolute error remaining below 0.016.
While other methods can approximate the general shape of the solution, they still exhibit significant prediction errors.
It should be noted that our operator model is trained at a resolution of $32\times 32\times32$ (for DeepONet-based models, details of the training data are provided in Table \ref{table-deeponetcnn} of the \ref{A-parameters}), extrapolation is performed at a higher resolution of $64\times 64\times64$.

\begin{table}[!ht]
\begin{center}\small
\caption{\small \rm Number of trainable parameters, size of training data,  times of training and extrapolation for each model relies on $\lambda$-$\omega$ RD systems, where the values before and after the slash ``/" represent the  size of training data for the branch net and trunk net respectively.}\label{table-cost}
\vspace{0.05in}
\begin{tabular}{ ccccc}
\hline
\hline\\[-2ex]
\rule{0pt}{10pt} \multirow{2}{*}{Model}  &  \multirow{2}{*}{No. of parameters}  &  \multirow{2}{*}{Size of training data}  & Time per & Time per step for \\[0.5ex]
\rule{0pt}{10pt}                         &                                      &                                  &                epoch (s)                     & extrapolation (s) \\[0.5ex]
\hline\\[-2ex]

\rule{0pt}{10pt}   DeepONet-MLP     &6.60M    & $(10,128^2)$/$(10000,2)$                & 0.26 & 0.0049     \\[0.5ex]

\rule{0pt}{10pt}   DeepONet-CNN     &1.66M  & $(5,2,128,128)$/$(10000,2)$               & 0.21 & 0.0049    \\[0.5ex]

\rule{0pt}{10pt}   FNO            &0.39M & $(5,128,128,4)$                               & 0.29 & 0.0019    \\[0.5ex]

\rule{0pt}{10pt}   ConvLSTM       &0.16M   & $(256,256,2)$                               & 1.35 & 0.0215    \\[0.5ex]

\rule{0pt}{10pt}   PIFNO           &0.39M   & $(5,128,128,4)$                            & 0.49 & 0.0019   \\[0.5ex]

\rule{0pt}{10pt}   PIdeepONet-CNN    &1.66M          & $(5,2,128,128)$/$(10000,2)$      & 0.95 & 0.0049   \\[0.5ex]

\hline
\hline\\[-2ex]
\end{tabular}
\end{center}
\end{table}

\section{Algorithm analysis}

\subsection{Computational cost}

Firstly, we take the $\lambda$-$\omega$ RD systems as a representative case to analyze the computational overhead of the PMNO based on different operator structures during the training and extrapolation stages. Table \ref{table-cost} compares the number of trainable parameters, size of training data, times of training and extrapolation for each model.
Considering that our study focuses on operator-based models, the number of trainable parameters is substantially larger compared to fixed-resolution methods ConvLSTM. In particular, due to the presence of two subnetworks, DeepONet-based models have significantly more parameters than FNO-based models. But the positional encoding in the trunk net reduces the dependence on high-resolution spatial input data, allowing DeepONet-based models to operate with small amounts of input data.
For instance, FNO-based models typically take input data of size $5\times128\times128\times4$, where spatial positions are inherently included. In comparison, DeepONet-CNN operates with significantly lower input volume, such as $5\times128\times128\times2+10000\times2$, since spatial locations are explicitly handled through the trunk network's positional encoding.
It should be noted that the CNN-based models in DeepONet, despite having fewer parameters owing to their shallow design, exhibit better performance than their MLP counterparts.
Importantly, the multi-step neural operator architecture introduces negligible additional memory overhead compared to single-step methods and allows for parallelized processing of individual time steps, significantly improving computational efficiency.

In terms of time cost, FNO-based models incur longer training time due to the computational overhead introduced by the integral kernel operators.
Moreover, training with physics guided loss functions also increases the training time during backpropagation, as it requires automatic differentiation to compute derivative information. However, this limitation is inherent to physics-informed learning approaches and is offset by the fact that such models require little to no training data.
Although neural operator predictors demand substantial training time, they offer much faster extrapolation than traditional numerical solvers, especially in high-dimensional problems.
This also highlights the motivation behind developing machine learning-based predictors.

\subsection{Sensitivity tests on major model parameters}

In the following, we investigate the impact of major hyperparameters on model performance.
The proposed PMNO method involves several hyperparameters, including the neural operator $\mathcal{G}$ architecture, the step size $k$ (the order of the time integrator), the number of recurrent iterations $L$, and among others. Notably, the step size $k$ serves as a key hyperparameter in our multi-step framework, substantially impacting model performance. In the previous section, we fixed $k=5$ and examined how different operator architectures affect predictive performance. Here, we focus on analyzing the influence of $k$ and the number of recurrent iterations $L$.

\begin{figure}[!t]
    \centering
  {\scalebox{0.7}[0.7]{\includegraphics{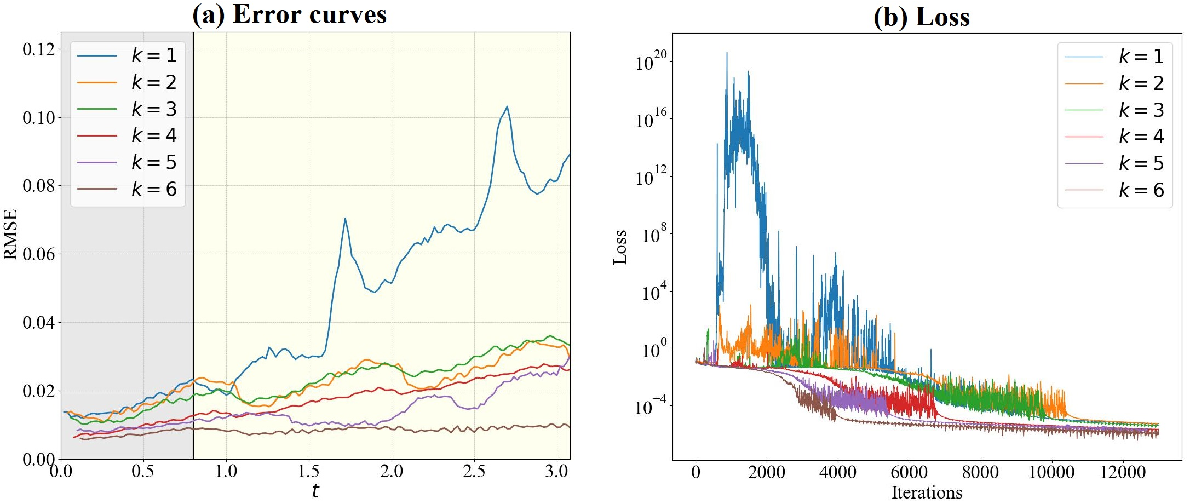}}}
  \vspace{-0.1in}
\caption{\small \rm Impact of $k$ on the performance of the PIFNO predictor for 2D advection equation. (a) Error propagation curves, where the gray region on the left represents the
training domain, while the yellow one on the right indicates the extrapolation domain. (b) The corresponding loss curves. }
\label{f-kadvection}
\end{figure}

{\bf The step size $k$:} The influence of the order $k$ on time integrator (\ref{bash}) has been extensively studied in numerical analysis \cite{num-ana}.
In general, the $k$-order method achieves $k$-th order accuracy. However, blindly increasing the order $k$ does not always lead to improved accuracy, as higher-order methods may violate stability conditions. For example, $k$-step BDFs are known to become unstable when $k>6$ (see \ref{A-BDF}).
In this test, we take the 2D advection equation as a representative case to compare the predictive accuracy for the PIFNO predictor with various step size, from $k=1$ to $k=6$.
Also with the time interval set to $\Delta t = 0.02$, the training time duration is $[0, 0.8]$.
The solution error propagation with time $\varepsilon(t)$ for different $k$ is displayed in Fig.~\ref{f-kadvection}(a).
It can be observed that the predictive performance of the network tends to improve with increasing values of $k$.
When $k=1$, the network structure essentially reduces to the spatial convolution combined with temporal recurrent architecture as described in \cite{universal-fno}.
Nevertheless, this type of network architecture exhibits limited predictive performance.
An increase in the order $k$ leads to a reduction in training error and, more importantly, a substantial improvement in the effective prediction horizon during the extrapolation phase.
Furthermore, an increasing order $k$ facilitates the training process, which is clearly demonstrated by the corresponding loss trajectories (see Fig.~\ref{f-kadvection}(b)).
For instance, single-step neural operator architectures tend to exhibit an initially large loss, with convergence occurring at relatively higher final loss values.
This may be attributed to the utilization of more past information for predicting the next state, thereby enhancing the prediction accuracy.
These discussions highlight the superior properties of the multi-step neural operator architecture, both during training and in the extrapolation phase.

\begin{figure}[!t]
    \centering
  {\scalebox{0.7}[0.7]{\includegraphics{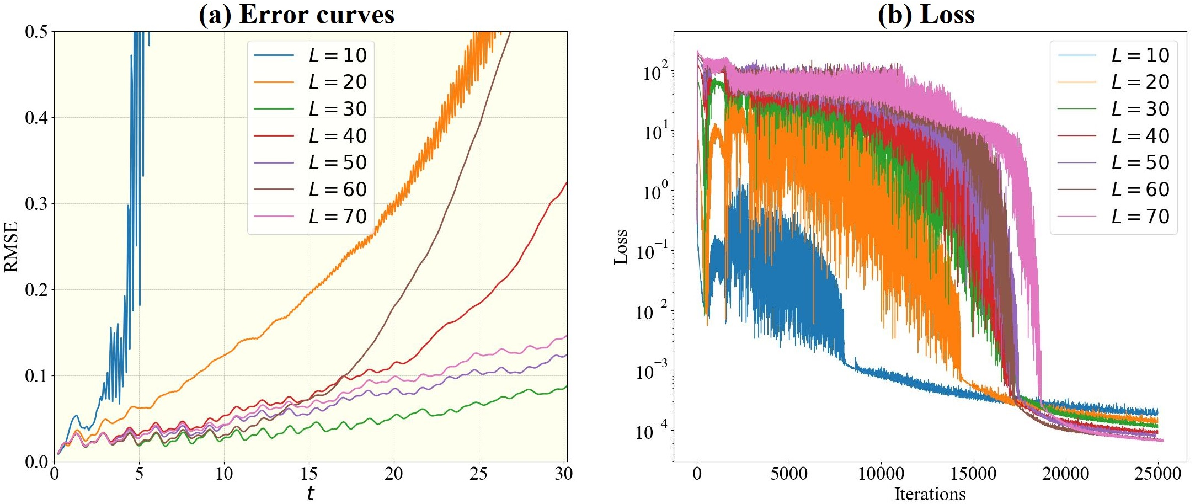}}}
  \vspace{-0.1in}
\caption{\small \rm Impact of $L$ on the performance of the PIdeepONet-CNN predictor for $\lambda$-$\omega$ RD equations. (a) Error propagation curves at $k=5$. (b) The corresponding loss curves. }
\label{f-LRD}
\end{figure}

{\bf The number of recurrent iterations $L$:} As another important factor influencing prediction performance, we evaluate the effect of training length $L$, i.e. the number of recurrent iterations, from $L=10$ to $L=70$, on the model's accuracy for PIdeepONet-CNN predictor using $\lambda$-$\omega$ RD equations as an example. We take $\Delta t=0.05$ and fix $k=5$. The solution error propagation with time $\varepsilon(t)$ for different $L$ is exhibited in Fig.~\ref{f-LRD}(a). As expected, a shorter training length $L$ prevents the model from effectively capturing the underlying physical dynamics of the system, resulting in a rapid increase in prediction error during the extrapolation phase (see the curves at $L=10$ and $L=20$ in Fig.~\ref{f-LRD}(a)).
However, a longer training period does not necessarily guarantee better prediction. Specifically, we observe that when $L=60$, although the model achieves low error in the short term, the prediction accuracy deteriorates significantly over longer time horizons. Interestingly, the model trained with $L=30$ achieves the best performance overall, even surpassing the cases with
$L=50$ and $L=60$. A longer training horizon typically corresponds to a deeper recurrent structure, which not only increases the training difficulty but also raises the risk of overfitting, ultimately hindering the model’s extrapolation capability. Fig.~\ref{f-LRD}(b) illustrates the loss curves during the training process. It can be observed that as the training length $L$ increases, the model generally requires a greater number of iterations for the loss to begin decreasing significantly.
These observations highlight the need to carefully balance the training duration to ensure sufficient learning while maintaining model stability and generalization.

\section{Conclusions and discussions}

Overall, in this paper we have proposed a novel neural operator architecture, the PMNO method, which incorporates the advantages of linear multi-step methods to address the challenge in long-horizon prediction of complex physical systems.
The PMNO framework combines the strengths of multi-step temporal modeling in forward pass and implicit time-stepping training scheme based on BDF during backpropagation, achieving state-of-the-art performance across challenging physical modeling tasks, including 2D linear system, modeling over irregular domain, complex-valued wave dynamics, and reaction-diffusion processes.
The key innovations of PMNO include: (a) the multi-step neural operator structures, (b) BDF guided training paradigm with fewer data samples, (c) the causal training strategy, (d) fast extrapolation on arbitrary spatial resolutions.
Compared to single-step approaches, it demonstrates enhanced accuracy and stability by leveraging information from multiple past time steps. This not only improves predictive performance over extended time horizons but also facilitates more efficient and stable training. The robustness and extrapolation capability observed across various tasks underscore its potential as a powerful framework for predicting complex dynamical systems.

Despite its promising performance, the proposed PMNO architecture can be further improved in multiple areas.
Firstly, the choice of the core operator component $\mathcal{G}$ significantly influences the predictive performance. Although various neural operator architectures, such as FNO, DeepONet, and their variants, can be seamlessly integrated into the PMNO framework, selecting the most suitable operator is often nontrivial and problem-dependent.
For instance, in the case of irregular spatial domains, integrating graph neural operator structures provides a natural and effective means of enhancing model performance by better capturing geometric and topological information \cite{GKN}. Secondly, DeepONet-based models often exhibit suboptimal performance in certain scenarios, such as when the target solution exhibits localized behavior or in high-dimensional settings (see Examples \ref{ex1} and \ref{ex5}). This limitation may stem from the Monte Carlo sampling strategy employed in the trunk net, which can lead to inadequate representation of localized or complex input features.
In fact, owing to the causal training strategy, the output solutions at earlier time steps are relatively accurate, which enables their effective utilization in designing adaptive learning mechanisms. Therefore, adaptive sampling strategies can be designed to effectively capture the spatial information such as the Neural Galerkin schemes introduced in Ref. \cite{neural-galerkin}.
Lastly,  training with physics-guided loss functions typically increases the training time, as automatic differentiation techniques exhibit exponential scaling in both memory and computation, especially when higher-order derivatives are involved. To address this drawback, some derivative estimators can be employed to accelerate the computation such as Taylor-mode automatic differentiation and Stochastic Taylor derivative estimator \cite{derivate1,derivate2}.

\v\v\v \v \noindent {\bf CRediT authorship contribution statement}

\v Jin Song: Conceptualization, Formal analysis, Methodology, Validation, Writing – original draft. Kenji Kawaguchi: Supervision, Writing – review \& editing. Zhenya Yan: Conceptualization, Methodology, Project administration, Supervision, Writing – review \& editing.

\v \v \noindent {\bf Declaration of competing interest}

\v The authors declare that they have no known competing financial interests or personal relationships that could have appeared to influence the work reported in this paper.

\v \v \noindent {\bf Data availability}

\v All the data and codes are available in GitHub at https://github.com/JinSong99/PMNO.

\v \v \noindent {\bf Acknowledgement}

\vspace{0.05in}
 This work was partially supported by the  National Key Research and Development Program of China (No. 2024YFA1013101) and the National Natural Science Foundation of China (No.12471242).

\appendix
\renewcommand{\thesection}{Appendix \Alph{section}}

\section{Neural operators} \label{A-NO}

\subsection{Deep operator network (DeepONet)\cite{deeponet}}
DeepONet consists of two sub-networks: a branch network and a trunk network, where the branch net takes the discretized function $\bu$ at sensors $\{\bx_1, \bx_2, \ldots, \bx_m\}$ as input, and trunk net encodes the coordinates $\bm y$ of the output function. Note that $\bx_i$ is independent of $\bm y$. Each of the two networks produces $p$ outputs, which are denoted as $[b_1, b_2, \ldots, b_p]^T$ and $[t_1, t_2, \ldots, t_p]^T$, respectively. Finally, the final output is derived by performing the dot product between the outputs of the two sub-networks with a bias:
\begin{equation}\label{deeponet-output}
  \mathcal{G}(\bu)(\bm y) = \frac{1}{p}\sum_{i=1}^{p}b_i(\bu)t_i(\bm y) + b_0,
\end{equation}
where $b_0\in\mathbb{R}$ is a bias. In this formulation, the branch network produces the coefficients, while the trunk network generates the basis functions of the target output function.
Moreover, the choice of the branch network is flexible, which can be designed as a fully connected network, residual network, convolutional neural network or a graph neural network depending on the structure of the input function.

\subsection{Fourier neural operator (FNO)\cite{fno}}
Fourier neural operator is defined in the following form,
\begin{equation}\label{fno}
  \mathcal{G}: = \mathcal{Q}\circ (\mathcal{W}_L+\mathcal{K}_L)\circ \cdots \circ \sigma(\mathcal{W}_1+\mathcal{K}_1) \circ \mathcal{P},
\end{equation}
where $\mathcal{P}$ and $\mathcal{Q}$ are pointwise operators that encode the lower dimension function into higher dimensional space and vice versa. And the model stacks $L$ layers of $\sigma(\mathcal{W}_l+\mathcal{K}_l)$ with pointwise linear operators $\mathcal{W}$ and integral kernel operators $\mathcal{K}$. $\sigma$ are fixed activation functions.
$\mathcal{P}$ and $\mathcal{Q}$ are parameterized with neural networks in the corresponding finite dimensional space, and $\mathcal{W}$ are parameterized as matrices.
The key mechanism of FNO lies in the use of Fourier transforms within the integral kernel operators $\mathcal{K}$. By applying the Fast Fourier Transform (FFT), the input is projected into the frequency domain, where low-frequency modes undergo a linear transformation and high-frequency modes are filtered out. Then the result is brought back to the spatial domain via an inverse FFT.
the Fourier integral operator is defined as for the input function $\bm v$,
\begin{equation}\label{fio}
  (\mathcal{K}\bm v)(\bx): = \mathcal{F}^{-1}\left(R \cdot (\mathcal{F} \bm v)\right)(\bx),
\end{equation}
where $\mathcal{F}$ and $\mathcal{F}^{-1}$ are the Fourier transform and its inverse, and $R$ is the parameters to be learned. Since the input functions are discretized during the forward process and the operators are defined pointwise, we incorporate the coordinate information into the input $[\bu, \bm y]$, where $\bu$ are the input functions and $\bm y$ are the coordinates of the output function.

\section{Backward Differentiation Formula}\label{A-BDF}
Backward Differentiation Formula (BDF) is an implicit linear multi-step method, particularly effective for stiff problems \cite{BDF}. Its core idea is to approximate derivatives using multi-step backward differences, thereby increasing the accuracy of the approximation.
Different from the general linear multi-step method, the BDF scheme is implicit since it involves only the information of current-step function $\mathcal{N}[\bx, t_{i+k}; \bu_{i+k}]$ on the right-hand side. It can be formulated as follows,
\begin{equation}\label{BDF}
  \sum_{j=0}^{k}a_j\bu_{i+j} = \Delta t b_k\mathcal{N}[\bx,t_{i+k};\bu_{i+k}], \quad i=0,1,\ldots.
\end{equation}
The $k$-step BDFs with $k\leq 6$ are stable \cite{a-stability}, and can be expressed in the following form,
\begin{equation}\label{1-bdf}
  k=1: \quad \bu_{i+1} - \bu_{i} = \Delta t\mathcal{N}[u_{i+1}],
\end{equation}

\begin{equation}\label{2-bdf}
  k=2: \quad \bu_{i+2} - \frac{4}{3}\bu_{i+1} + \frac{1}{3}\bu_{i} = \frac{2}{3}\Delta t\mathcal{N}[\bu_{i+2}],
\end{equation}

\begin{equation}\label{3-bdf}
  k=3: \quad   \bu_{i+3} - \frac{18}{11}\bu_{i+2} + \frac{9}{11}\bu_{i+1} - \frac{2}{11}\bu_{i} = \frac{6}{11}\Delta t\mathcal{N}[\bu_{i+3}],
\end{equation}

\begin{equation}\label{4-bdf}
  k=4: \quad   \bu_{i+4} - \frac{48}{25}\bu_{i+3} + \frac{36}{25}\bu_{i+2} - \frac{16}{25}\bu_{i+1} + \frac{3}{25}\bu_{i}= \frac{12}{25}\Delta t\mathcal{N}[\bu_{i+4}],
\end{equation}

\begin{equation}\label{5-bdf}
  k=5: \quad   \bu_{i+5} - \frac{300}{137}\bu_{i+4} + \frac{300}{137}\bu_{i+3} - \frac{200}{137}\bu_{i+2} + \frac{75}{137}\bu_{i+1} - \frac{12}{137}\bu_{i}= \frac{60}{137}\Delta t\mathcal{N}[\bu_{i+5}],
\end{equation}

\begin{equation}\label{6-bdf}
  k=6: \quad   \bu_{i+6} - \frac{360}{147}\bu_{i+5} + \frac{450}{147}\bu_{i+4} - \frac{400}{147}\bu_{i+3} + \frac{225}{147}\bu_{i+2} - \frac{72}{147}\bu_{i+1} + \frac{10}{147}\bu_{i}= \frac{60}{147}\Delta t\mathcal{N}[\bu_{i+6}].
\end{equation}

\section{The network structures, hyperparameters and model parameters}\label{A-parameters}

We list the configuration of parameters across different physical systems in Table \ref{table-parameter}.
Meanwhile, the summary of each network architectures and hyperparameters for different physical systems is exhibited in Tables \ref{table-deeponetcnn}, \ref{table-deeponetmlp}, \ref{table-fno}, and  \ref{table-convlstm}. Besides, Xavier \cite{xavier} weight initialization is used in this paper.

\begin{table}[!t]
\vspace{0.05in}
\begin{center}\small
\caption{\small \rm Configuration of parameters across different physical systems, where $N$ denotes the number of sample points for DeepONet-based models.}\label{table-parameter}
\vspace{0.05in}
\begin{tabular}{ ccccc}
\hline
\hline\\[-2ex]

\rule{0pt}{10pt}    Systems       & $k$    & $\Delta t$    & $L$  & $N$   \\[0.5ex]
\hline\\[-2ex]
\rule{0pt}{10pt}    Advection              & 5      & 0.02 & 36 & 3000    \\[0.5ex]
\rule{0pt}{10pt}    $\lambda$-$\omega$ RD  & 5      & 0.05 & 46 & 10000   \\[0.5ex]
\rule{0pt}{10pt}    Heat                   & 5      & 0.03 & 36 & 5000    \\[0.5ex]
\rule{0pt}{10pt}    NLS                    & 5      & 0.15 & 31 & 25000   \\[0.5ex]
\rule{0pt}{10pt}    GS                     & 5      & 0.4  & 41 & 18000   \\[0.5ex]

\hline
\hline
\end{tabular}
\end{center}
\end{table}

\begin{table}[!ht]
\vspace{0.05in}
\begin{center}\small
\caption{\small \rm Summary of DeepONet-CNN architectures and hyperparameters for different physical systems,
where the kernel size of all convolutions (Conv2d and Conv3d) is set to 5, and zero-padding is applied to preserve the spatial dimensions of the data.}\label{table-deeponetcnn}
\vspace{0.05in}
\begin{tabular}{ ccccc}
\Xhline{1pt}
\rule{0pt}{10pt}         Systems                &     &  Size of input  & Structures   & Size of output  \\[0.5ex]
\Xhline{1pt}

\rule{0pt}{10pt}   \multirow{2}{*}{Advection}  &   Branch net      & $(5,1,128,128)$      & [Conv2d, Tanh]+[$100^2$, 100] & \multirow{2}{*}{$(3000,1)$}   \\[0.5ex]
\rule{0pt}{10pt}                               &   Trunk net       & $(3000,2)$           & [2, 100, Tanh, 100, Tanh, 100]    &    \\[0.5ex]
\hline

\rule{0pt}{10pt}   \multirow{2}{*}{$\lambda$-$\omega$ RD}  &   Branch net      & $(5,2,128,128)$      & [Conv2d, Sin]+[$128^2$, 100] & \multirow{2}{*}{$(10000,2)$}   \\[0.5ex]
\rule{0pt}{10pt}                                           &   Trunk net       & $(10000,2)$          & [2, 100, Sin, 100, Sin, 100]    &    \\[0.5ex]
\hline

\rule{0pt}{10pt}   \multirow{2}{*}{Heat}       &   Branch net      & $(5,1,256,256)$      & [Conv2d, Tanh]+[$256^2$, 100] & \multirow{2}{*}{$(5000,1)$}   \\[0.5ex]
\rule{0pt}{10pt}                               &   Trunk net       & $(5000,2)$           & [2, 100, Tanh, 100, Tanh, 100, Tanh, 100]    &    \\[0.5ex]
\hline

\rule{0pt}{10pt}   \multirow{2}{*}{NLS}        &   Branch net      & $(5,2,32,32,32)$      & [Conv3d, Tanh]+[$32^3$, 100] & \multirow{2}{*}{$(25000,2)$}   \\[0.5ex]
\rule{0pt}{10pt}                               &   Trunk net       & $(25000,3)$           & [3, 100, Tanh, 100, Tanh, 100]    &    \\[0.5ex]
\hline

\rule{0pt}{10pt}   \multirow{2}{*}{GS}         &   Branch net      & $(5,2,32,32,32)$      & [Conv3d, Tanh]+[$32^3$, 100] & \multirow{2}{*}{$(18000,2)$}   \\[0.5ex]
\rule{0pt}{10pt}                               &   Trunk net       & $(18000,3)$           & [2, 100, Tanh, 100, Tanh, 100]    &    \\[0.5ex]

\Xhline{1pt}
\end{tabular}
\end{center}
\end{table}

\begin{table}[!ht]
\vspace{0.05in}
\begin{center}\small
\caption{\small \rm Summary of DeepONet-MLP architectures and hyperparameters for different physical systems.}\label{table-deeponetmlp}
\vspace{0.05in}
\begin{tabular}{ ccccc}
\Xhline{1pt}
\rule{0pt}{10pt}         Systems                &     &  Size of input  & Structures   & Size of output  \\[0.5ex]
\Xhline{1pt}

\rule{0pt}{10pt}   \multirow{2}{*}{$\lambda$-$\omega$ RD}  &   Branch net      & $(5\times 2,128^2)$      & [$128^2$, 100, Sin, 100, Sin, 100] & \multirow{2}{*}{$(10000,2)$}   \\[0.5ex]
\rule{0pt}{10pt}                                           &   Trunk net       & $(10000,2)$          & [2, 100, Sin, 100, Sin, 100]    &    \\[0.5ex]
\hline

\rule{0pt}{10pt}   \multirow{2}{*}{Heat}       &   Branch net      & $(5,5000)$      & [5000, 60, Tanh, 60, Tanh, 60, Tanh, 10] & \multirow{2}{*}{$(5000,1)$}   \\[0.5ex]
\rule{0pt}{10pt}                               &   Trunk net       & $(5000,2)$           & [2, 60, Tanh, 60, Tanh, 60, Tanh, 10]    &    \\[0.5ex]

\Xhline{1pt}
\end{tabular}
\end{center}
\end{table}

\begin{table}[!ht]
\vspace{0.05in}
\begin{center}\small
\caption{\small \rm Summary of FNO architectures and hyperparameters for different physical systems,
where $\mathcal{K}$ is integral kernel operator parameterized by Fourier transforms with different Fourier modes,
the filter size of all convolutions (Conv2d and Conv3d) is set to 1, and the channels are set as width.}\label{table-fno}
\vspace{0.05in}
\begin{tabular}{ cccccc}
\hline
\hline\\[-2ex]
\rule{0pt}{10pt}         Systems                    &  Size of input  & Structures  & Modes & Width & Size of output  \\[0.5ex]
\hline\\[-2ex]

\rule{0pt}{10pt}   Advection        & $(5,128,128,3)$      & [3, 32]+$3\times$[$\mathcal{K}$, Conv2d]+[32, 64, GLUE, 1] & 8  & 32 & $(128, 128, 1)$   \\[0.5ex]

\rule{0pt}{10pt}   $\lambda$-$\omega$ RD       & $(5,128,128,4)$      & [4, 32]+$3\times$[$\mathcal{K}$, Conv2d]+[32, 64, GLUE, 2] & 8 & 32 & $(128, 128, 2)$   \\[0.5ex]

\rule{0pt}{10pt}   Heat             & $(5,256,256,3)$      & [3, 64]+$3\times$[$\mathcal{K}$, Conv2d]+[64, 64, GLUE, 1] & 16 & 64 & $(256, 256, 1)$   \\[0.5ex]

\rule{0pt}{10pt}   NLS          & $(5,32, 32, 32, 4)$      & [4, 32]+$3\times$[$\mathcal{K}$, Conv3d]+[32, 64, GLUE, 2] & 16 & 32 & $(32, 32, 32, 2)$   \\[0.5ex]

\rule{0pt}{10pt}   GS              & $(5, 32, 32, 32, 4)$      & [4, 32]+$3\times$[$\mathcal{K}$, Conv3d]+[32, 32, GLUE, 2] & 8 & 32 & $(32, 32, 32, 2)$   \\[0.5ex]

\hline
\hline\\[-2ex]
\end{tabular}
\end{center}
\end{table}

\begin{table}[!ht]
\vspace{0.05in}
\begin{center}\small
\caption{\small \rm Summary of ConvLSTM architectures and hyperparameters for different physical systems.}\label{table-convlstm}
\vspace{0.05in}
\begin{tabular}{ ccccc}
\hline
\hline\\[-2ex]
\rule{0pt}{10pt}         Systems                    &  Size of data  & Filter size  & Layers & Channels  \\[0.5ex]
\hline\\[-2ex]

\rule{0pt}{10pt}   Advection        & $(128,128,1)$      & 5 & 2  & 32   \\[0.5ex]

\rule{0pt}{10pt}   $\lambda$-$\omega$ RD       & $(256,256,2)$      & 5 & 2 & 32   \\[0.5ex]

\rule{0pt}{10pt}   Heat             & $(256,256,1)$      & 5 & 2 & 32  \\[0.5ex]

\rule{0pt}{10pt}   NLS          & $(32, 32, 32, 2)$      & 5 & 1 & 16    \\[0.5ex]

\rule{0pt}{10pt}   GS              & $(32, 32, 32, 2)$      & 5 & 1 & 16   \\[0.5ex]

\hline
\hline\\[-2ex]
\end{tabular}
\end{center}
\end{table}

\end{document}